\documentclass[journal]{IEEEtai}
\usepackage[colorlinks,urlcolor=blue,linkcolor=blue,citecolor=blue]{hyperref}
\ifCLASSINFOpdf
\else
\fi
\ifCLASSINFOpdf
   \usepackage[pdftex]{graphicx}
   \graphicspath{{../Figures/}}
   \DeclareGraphicsExtensions{.pdf,.jpeg,.png}
\else
\fi
\usepackage{amsmath}
\usepackage{amssymb}
\usepackage{comment}

\usepackage{amsmath}
\usepackage{amsmath,amsfonts}

\usepackage{amsmath,amsfonts}
\usepackage{algorithmic}
\usepackage{array}
\usepackage[caption=false,font=footnotesize]{subfig}
\usepackage{textcomp}
\usepackage{stfloats}
\usepackage{url}
\usepackage{verbatim}
\usepackage{graphicx}
\usepackage[table]{xcolor}
\usepackage[utf8x]{inputenc}
\usepackage{stfloats}
\usepackage{tabularx,booktabs}
\usepackage{multirow}
\usepackage{amssymb}
\usepackage{array}
\usepackage{xcolor}
\usepackage{cite}
\usepackage{array}

\ifCLASSOPTIONcompsoc
 \usepackage[caption=false,font=normalsize,labelfont=sf,textfont=sf]{subfig}
\else
  \usepackage[caption=false,font=footnotesize]{subfig}
\fi

\usepackage{stfloats}
\usepackage{tabularx,booktabs}

\usepackage{algorithm2e}

\usepackage{multirow}

\usepackage{amssymb}
\usepackage{array}
\usepackage{xcolor}
\usepackage{cite}
\begin{document}
\title{RAFA-Net: Region Attention Network For Food Items And Agricultural Stress Recognition}

\author{Asish~Bera,~\IEEEmembership{Member,~IEEE}, 
Ondrej Krejcar, and
        Debotosh~Bhattacharjee,~\IEEEmembership{Senior Member,~IEEE}
    
\thanks{A. Bera is with the Department of Computer Science and Information Systems, Birla Institute of Technology and Science, Pilani, Pilani Campus,  Rajasthan, 333031, India.  E-mail: asish.bera@pilani.bits-pilani.ac.in}
\thanks{O. Krejcar is with the Center for Basic and Applied Science, Faculty of Informatics and Management, University of Hradec Kralove, Rokitanskeho 62, 500 03 Hradec Kralove, Czech Republic. He is also with the   Skoda Auto University, Na Karmeli 1457, 293 01 Mlada Boleslav, Czech Republic, and the Malaysia Japan International Institute of Technology (MJIIT), Universiti Teknologi Malaysia, Kuala Lumpur, Malaysia. Email: ondrej.krejcar@uhk.cz}
\thanks{D. Bhattacharjee is with the Department of Computer Science and Engineering, Jadavpur University, Kolkata 32, India.  He is also with the Center
for Basic and Applied Science, Faculty of Informatics and Management,
University of Hradec Kralove, Rokitanskeho 62, 500 03 Hradec Kralove,
Czech Republic.  Email:  debotosh@ieee.org}

}

\markboth{IEEE Transactions on AgriFood Electronics, Vol. 00, No. 0,  September~2024}%
{RAFA-Net: Region Attention Network For Food And Agricultural Stress  Recognition}

\maketitle
\author{ } 
\begin{abstract}
Deep Convolutional Neural Networks (CNNs) have facilitated remarkable success in recognizing various food items and agricultural stress. A decent performance boost has been witnessed in solving the agro-food challenges  by mining and analyzing of region-based partial feature descriptors. Also, computationally expensive ensemble learning schemes fusing multiple CNNs  have been studied in earlier works. This work proposes  a region attention scheme  for modeling long-range dependencies by building a correlation among different regions within an input image. The attention method enhances feature representation by learning the usefulness of context information from complementary regions. Spatial pyramidal pooling and  average pooling pair aggregate  partial descriptors into a holistic representation. Both pooling methods establish  spatial and channel-wise relationships  without incurring extra parameters. A context gating scheme is applied to refine the descriptiveness of weighted attentional features, which is relevant for classification. The proposed Region Attention network for Food items and Agricultural stress recognition method, dubbed RAFA-Net, has been experimented on three  public food datasets, and has achieved  state-of-the-art performances with distinct margins. The highest top-1 accuracies of RAFA-Net are 91.69\%, 91.56\%, and 96.97\% on  the UECFood-100, UECFood-256, and MAFood-121 datasets, respectively. In addition, better accuracies have been achieved on two  benchmark agricultural stress datasets. The best top-1 accuracies on the Insect Pest (IP-102) and PlantDoc-27 plant disease datasets are 92.36\%, and 85.54\%, respectively;  implying RAFA-Net's generalization capability.

\end{abstract}

\begin{IEEEkeywords}
 Food Items, Agricultural Stress, Plant Disease, Insect Pest,  Feed-Forward Network, Region Attention.
\end{IEEEkeywords}

%
\IEEEpeerreviewmaketitle

\section{Introduction}
\IEEEPARstart{F}{ood} industry encounters a revolutionary demand for food-supply and safety due to rising  population  and climate change around the world. Environmental changes play a crucial role in maintaining  the ecosystem and  food production. Mainly,  automation in food technology, nutrition estimation, dietary control, ingredient recognition,  and  food processing  applications using machine learning (ML) techniques have been studied in the recent past \cite{rayhana2023review, wang2022ingredient, arslan2021fine}.  Food production is strongly related to crop yield, which  is heavily influenced by plant diseases and harmful insect pests. Plant diseases and insect pests hinder crop yield and  cause biotic plant stress, named here agricultural stress.  Automated plant health monitoring and  detection of plant stress  using  ML is vital for agriculture \cite{donapati2023real}. ML algorithms are beneficial than  labor-intensive manual techniques to combat these challenges.  Though, decent progress  of automation in agro-food sectors leveraging ML  has been witnessed in recent time, yet, ample research attention is needed for  socioeconomic growth and industrial readiness. We have united both challenges of agro-centric food processing cycle based on similitude in textural appearances,  and cohered for sustainable societal demands, especially, recognition of food dishes and plant stress.
 Here, an open concern is that can we develop a generalized ML solution to harness both recognition tasks? As a solution, this work proposes a generic deep learning method tailored for the classifications of food items and plant stress.

Food items exhibit a complex spatial arrangement due to  variations in lighting, color, shape, texture, and view \cite{yanai2015food}. Likewise, classifying infected plant leaves with  discernible  appearance variations in an intricate field environment is challenging.  These contextual disparities lead to a higher intra-class differences which cause difficulties in  recognition. Several  ML techniques have proven promising success in addressing these challenges of food and agriculture sectors. 

Convolutional neural networks (CNNs) have boosted the performances of several visual recognition challenges  related to food technology and precision agriculture  significantly
\cite{rayhana2023review}, \cite{konstantakopoulos2023review}. CNNs effectively learn the object's structural details and correlated information  via successive non-linear transformations of  convoluted feature maps. The global image-level description and local region-level contexts have proven their essentiality  by mining fine details for solving diverse visual recognition tasks. For example, some works have been devised by exploring visual attention \cite{bera2021attend,liang2020mvanet},  
advanced image augmentation \cite{zhong2020random, bera2023fine}, ensemble learning, lightweight deep architectures \cite{sheng2024lightweight}, etc. These kinds of vision-based deep models are suitable for the development of agricultural and food processing robots built with the cognitive and decision-making modules \cite{raja2024software}.  

The \textbf{R}egion \textbf{A}ttention Network for \textbf{F}ood items and \textbf{A}gricultural stress recognition, dubbed RAFA-Net, is  proposed  by correlating and mining region-based attentional information into a holistic feature map.  As food items are inherently comprised of complex patterns, neighboring context information is crucial for categorization. The proposed RAFA-Net  integrates  region-based  information into a precise feature map  for  classification.  The aim is to formulate  a semantic correlation among  partial descriptors by attending image regions to define a comprehensive feature vector. To achieve this, a deep feature map is computed from an input image leveraging a base CNN. The output feature map comprises a smaller spatial dimension based on which establishing spatial relation among discriminative regions is cumbersome. For this intent, the feature map is up-scaled into a higher spatial size for computing and correlating region-level description.

The attention mechanism has showcased its suitability in building long-distant dependency modeling for image classification and serves as a key component of diverse deep learning (DL) architectures \cite{vaswani2017attention}.  Due to its superior performance, a wide variety of self-attention regimes have  been developed. 
Intuitively, it is hypothesized that self-attention is beneficial when included at later layers within a deep network due to its adaptive nature in capturing finer details with arbitrary receptive field size, \textit{i.e.}, long-distant dependency modeling. It is witnessed  that  visual attention could improve  performance when included with a base CNN, which is followed here. 

On the other side, separable convolutions, parsimoniously designed convolutional architecture with low-rank factorization power, capture local-contexts by decoupling spatial and channel-wise information for building a correlation in the feature space \cite{chollet2017xception}. Typically, this blending offers a better representation power for high-level aggregation at a later stage of a CNN. These important aspects play a  key-role for designing the proposed RAFA-Net model.

In addition,  pyramidal pooling summarizes local-contexts at multiple granularity towards  deeper layers of a network, generally, before a fully connected layer. However, its superiority can also be exploited within interim layers, while context-aware feature learning is essential for aggregation at multi-scale.  
The spatial pyramid pooling (SPP) layer pools and aggregates features at a deeper layer allowing arbitrary aspect ratios. Finally, SPP generates fixed-length output vectors for multi-level pyramidal bins \cite{he2015spatial}. 

\begin{figure}[h]
\centering
\includegraphics[width=0.99\linewidth, ]{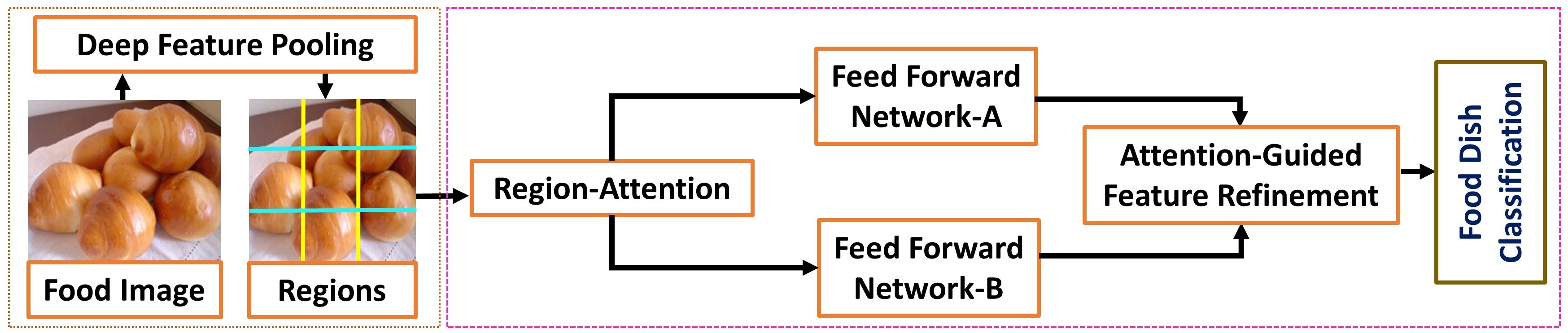}
\caption{The proposed RAFA-Net consists of a region-pooling module upon which attention is applied for enhancing distinctness of various regions. The spatial and channel-wise relationships among attentioned regions are captured by feed-forward paths, and the features are further refined for classification. }
\label{fig:Model1a}
\vspace{ -0.4 cm}
\end{figure}

Here, both pyramidal pooling and region-level average pooling strategies  have been explored through feed-forward networks (FFNs)  for aggregating deep features efficiently, denoted as $FFN_A$ and $FFN_B$, respectively.  
The $FFN_A$  aggregates local spatial information by  pooling into multi-level spatial bins in a hierarchical fashion. The responsiveness of $FFN_A$  determines  object's local distinctness into fixed-length descriptors. In $FFN_B$, average-pooling aggregates overall spatial details into a holistic feature map, and contributes for rendering channel-wise attention map and attention weights beyond. Thus, $FFN_B$ summarizes global informativeness of the descriptors. Afterward, with successive feature transformations via both FFNs,  the local ($FFN_A$) and global ($FFN_B$) feature vectors are combined to produce an attention-guided descriptor.  The  RAFA-Net  aggregates complementary description by mining visual distinctness of image-regions into an efficient feature map. A high-level representation  showcasing  the  key modules of proposed method is depicted in Fig. \ref{fig:Model1a}.

The RAFA-Net outperforms existing methods  tested on three  public image datasets of food items: UECFood-100 \cite{matsuda2012recognition}, UECFood-256 \cite{kawano2015automatic}, and MAFood-121  \cite{aguilar2019regularized}. Two  benchmark image datasets representing insect pests (IP-102 \cite{wu2019ip102}) and plant diseases (PlantDoc-27 \cite{singh2020plantdoc}) have been tested for agricultural stress recognition. This work contributes the following:

\begin{itemize}
    \item The region attention mechanism enhances the relevance  of partial feature descriptors in classifying images of food items and agricultural stress.
    
    \item The fusion of two pooling strategies through  feed-forward networks  calibrates  attentional feature refinement. 
    
    \item Experiments have been carried out on three public food datasets and two plant stress recognition datasets. The proposed RAFA-Net  has achieved state-of-the-art performances on those five datasets.  
\end{itemize}

The rest of this paper is organized as follows: Section \ref{rel_work}
summarizes related works. The proposed methodology is explained in Section \ref{method}.  Section \ref{experiments} demonstrates  experimental results and analysis. The conclusion is given in Section \ref{con}.
\vspace{ -0.2 cm}
\section{Related Works} \label{rel_work} 
Several methods have been developed for food classification and plant stress especially,  disease and insect pest classification, using hand-crafted feature computation (\textit{e.g.}, bag-of-words, keypoints descriptors, textures, etc.),  deeper/wider networks leveraging attention, ensemble techniques, and others.  A  study on recent  approaches in both areas is briefed here.

\vspace{ -0.3 cm}
\subsection{Food Item Classification}
The uncertainty-aware active learning framework considers the epistemic uncertainty. A thresholding criterion is imposed in epistemic uncertainty predictions to improve  accuracy \cite{pillai2022integrated}. The forward step-wise uncertainty-aware model selection (FS-UAMS) method proposed an ensemble  of CNNs  for food classification \cite{aguilar2022uncertainty}. 
A dataset consisting of 70 food classes in Qatar and the Middle East region is developed, and tested using a mobile-based deep CNN \cite{ansari2023mefood}. %
 An ensemble of two CNNs  is presented for a smartphone-based food recognition system \cite{fakhrou2021smartphone}. 
The joint-learning distilled network (JDNet) has developed a  multi-stage knowledge distillation  and instance activation learning  framework for training a student and  teacher networks simultaneously \cite{zhao2020jdnet}. A progressive self-distillation method has enhanced learning capacity by mining informative regions, apposite for food recognition \cite{zhu2023learn}.  
A multi-scale multi-view feature aggregation method unifies high-level semantics, mid-level attributes, and deep visual features  for food categorization \cite{jiang2019multi}. A joint learning method has explored ingredient-based semantic-visual graphs for food recognition and ingredient prediction \cite{wang2022ingredient}.  
Keypoints-based  region formulation and refined through attention have improved the visual recognition performance of food dishes \cite{bera2021attend}. An attention-based fusion has applied  multi-view attention for extracting and combining multiple semantic information from different tasks \cite{liang2020mvanet}.  A region-level attention network, comprising  a region-weighted module, is tested for food and ingredient joint classification \cite{xue2021region}.
Multi-dish recognition has explored  contextual relations from region features and dish labels \cite{deng2021mixed}.  A lightweight  global shuffle network (GSNet)  has been built to capture  long-range pixel dependencies  for recognizing food items \cite{sheng2024lightweight}.
Similar  CNNs and attention-based methods have   been studied related to agricultural stress. 
\vspace{ -0.3 cm}

\subsection{Plant Disease Classification}
Attention methods has become a key component of vision-based models. 
A multi-granular feature aggregation using the self-attention method is tested for crop disease classification \cite{zuo2022multi}. A lightweight  double fusion block with a coordinate attention module (DFCANet) is presented for corn leaf diseases in a real field environment  \cite{chen2022dfcanet}. A method exploring shuffle attention and HardSwish function is developed  for  tomato leaf disease recognition \cite{zhang2023ibsa_net}.   
A cross-attention module fitted to a bidirectional transposed feature pyramid network  is described for apple disease detection \cite{zhang2023precise}.
Multi-head self-attention  is used in a transformer model for  Cassava leaf disease detection \cite{thai2023formerleaf}. Trilinear CNN has been studied  for crop and disease identification \cite{wang2021t}. Recently, a lightweight CNN has been developed for classifying  rice plant diseases and experimented with a new dataset created in  real-field environment \cite{pal2024robust}. A  modified ResNet architecture supported with squeeze-and-excitation modules has been tested for strawberry disease detection \cite{wu2024strawberry}. A graph convolutional network (PND-Net) has been studied for plant stress classification \cite{bera2024pnd}.
A DL method comprising with residual blocks and self-attention (RBNet) has been tested with  dragonfly optimization for crop disease detection \cite{haider2024crops}. 
 The Inception Convolutional Vision Transformer (ICVT) has employed multi-head attention to emphasize discriminative regions for plant disease identification \cite{yu2023inception}
The generative adversarial network (GAN) has been utilized to generate data diversity from various small-scale  plant datasets \cite{cap2020leafgan}. A GAN has been developed  to improve the performance of grape leaf disease classification using local spot area  augmentation \cite{zhou2021grape}. Aligned with this line of study, we  have developed a generalized region based attention, enriched with multi-scale local-contexts for plant disease recognition in real-field conditions.

\vspace{ -0.1 cm}
\subsection{ Insect Pest Classification}
Another allied task to improve crop yield is pest control, which is a severe challenge in agriculture. Several automated methods have been developed for pest classification applying ML techniques. A deep learning (DL) method using a multi-scale attention  network has been presented for pest recognition \cite{feng2022ms}.  A lightweight ResNet8 model has been tested for few-shot pest classification \cite{guo2024lightweight}. The Insect Foundation model \cite{nguyen2024insect} has contributed a larger insect dataset and evaluated using a vision transformer (ViT) as backbone. Also, ensemble techniques and ViTs \cite{peng2022cnn}, and multi-branch  multi-scale attention method have been studied for pest control  \cite{ung2021efficient}. Recently, a dilated windows based ViT with efficient-suppressive self-attention (DWViT-ES) method has been developed \cite{hechen2024dilated}. 

\begin{figure}[h]
\centering
\includegraphics[width=0.32\linewidth, height=1.98 cm ]{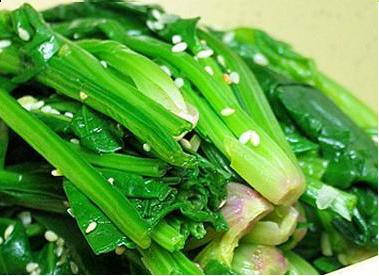} \hfill
\includegraphics[width=0.32\linewidth, height=1.98cm  ]{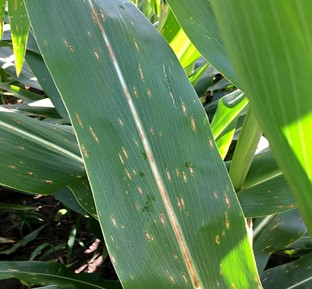} \hfill
\includegraphics[width=0.32\linewidth, height=1.98 cm  ]{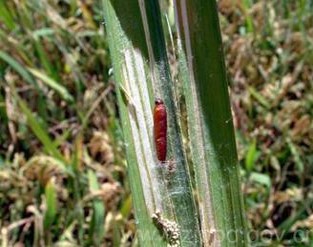}\hfill 
\vspace{-0.1 cm}
\subfloat[food-item]{\includegraphics[width=0.32\linewidth, height=1.98 cm  ]{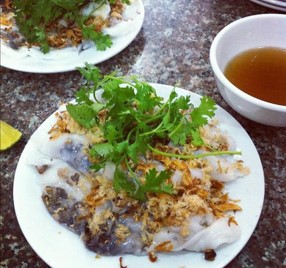}} \hfill
\subfloat[plant health]{\includegraphics[width=0.32\linewidth, height=1.98 cm  ]{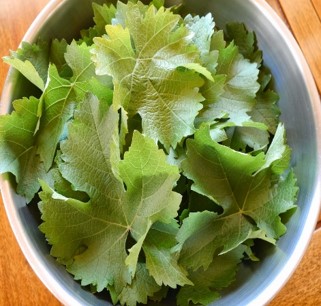}} \hfill
\subfloat[insect-pest]{\includegraphics[width=0.32\linewidth, height=1.98 cm  ]{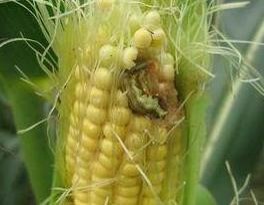}} \hfill
\caption{Visual variations in samples of different datasets, shown column-wise. Top row shows visual similarities among (a) food dish,  (b) plant disease, and (c) insect pest classes. Bottom row reflects their structural and texture variances in natural environment, implies  underlying recognition challenges.} 
\label{fig:DB1a}
\vspace{ -0.5 cm}
\end{figure}

\subsection{Research Gap Summary and Problem Formulation }

Existing works have recognized  either food items  or plant stress using DL methods.
A few ensemble of DL architectures have focused on performance gains and ignored the computational requirements  \cite{liang2020mvanet,arslan2021fine,konstantakopoulos2023review}. Whereas,  several lightweight CNNs have been designed to solve one problem,  which often fail to achieve competitive performances on similar other datasets \cite{liu2020deep}. Sometimes, precise adjustments are required in DL methods \textit{e.g.}, training with additional dataset \cite{pal2024robust}, etc.  Due to limited capacity of those shallower networks, experiments with complex datasets having hundreds of classes exhibit poor recognition performances. Thus, more sophisticated CNN  that builds correlation with global and local feature-level description is crucial. To address the limitations, we integrate two related problems \textit{i.e.}, food items and plant stress classifications into a unified challenge, and propose a common solution.

\noindent\textit{Hypothesis}:
A hypothesis in \cite{peng2022impact} suggests  that intensification of agricultural mechanization  through technological development including crop health monitoring and pest control can improve crop yield, thereby directing economic gains of the farmers. The hypotheses  in \cite{wang2023research}  imply that facility condition (\textit{i.e.}, the degree of support provided to a user for utilizing new technology), and adaptation intention of the farmers indicate a positive direction of promoting modernization in digital agriculture. However, due to  economic barrier of rural farmers, cost-effectiveness of a solution is of prime concern. Existing farming technology is crop or application specific, \textit{e.g.}, rice disease prediction. So, a  reusable method should be available to the farmers for different crops at various weather and field conditions.  To bridge this gap,  computational cost optimization of various agro-food recognition is addressed here. 

We hypothesize that  food items and agricultural stress classification problems exhibit similar structural layouts in their image-level descriptions (Fig. \ref{fig:DB1a}).  Row-wise inter-dataset samples show fair similarities, and column-wise examples exhibit wider variations, evident in Fig. \ref{fig:DB1a}. Precisely,  the variations in color, texture, lighting  and intricate background conditions in both agro-food recognitions  are  very similar and subtle in nature. So, an efficient technique is required for devising a common solution to accelerate economic gain.  Intuitively, to justify our hypothesis,  the RAFA-Net is proposed to widen its  effectiveness and applicability in agro-food classification. 

\begin{figure*}[h]
\centering
\subfloat{ 
\includegraphics[width=0.95\linewidth, ]{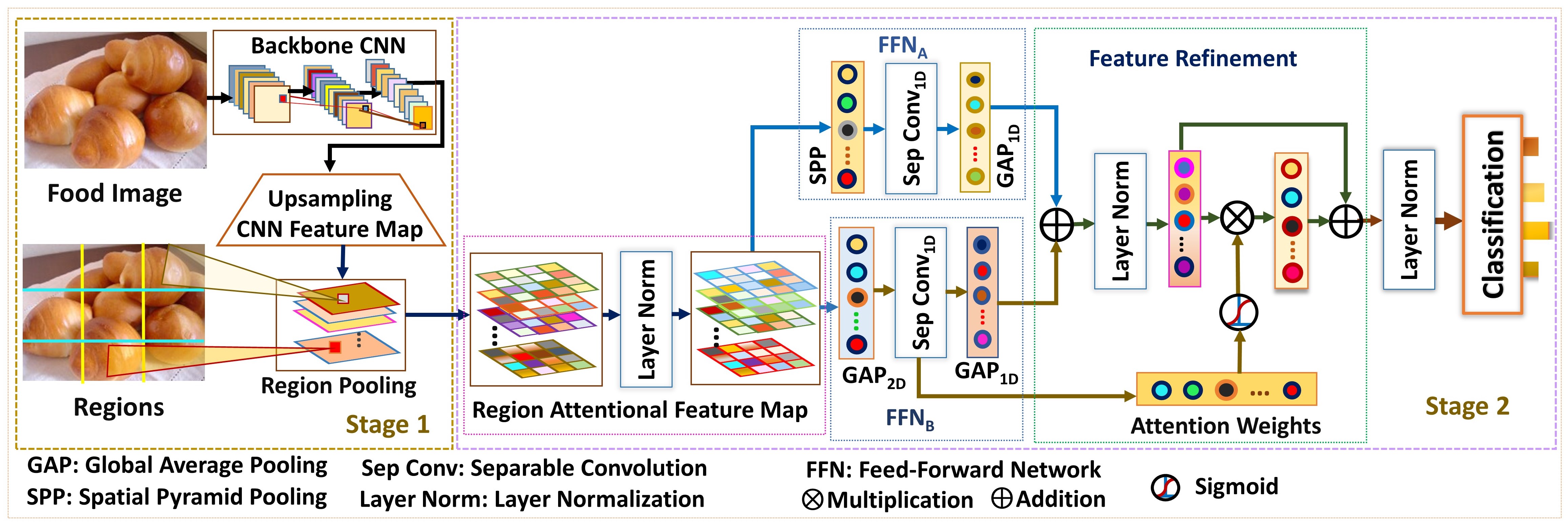}
}
\vspace{ -0.2 cm}
\caption{The proposed RAFA-Net consists of two sub-stages. a) Feature map pooling  from regions of the same size. b) Region attention technique and further refinement  by the feed-forward networks in two paths, which are combined together based on attention weights before the classification layer.}
\label{fig:Model2a}
\vspace{ -0.3 cm}
\end{figure*}

As said earlier, the RAFA-Net employs region attention for selecting discriminative features and  refines by attention weights through multi-scale pooling via feed-forward paths in its architecture. This generalized DL method boosts the performances of  both food items and plant stress recognition, justifying it's aptness  through rigorous experiments.

\section{Methodology} \label{method}
Our DL method  captures finer details of complex patterns  in an end-to-end trainable manner  using class labels. The proposed RAFA-Net, shown in Fig. \ref{fig:Model2a}, is comprised of two key stages: 1) deriving non-overlapped and fixed-size regions from the output feature map of a base CNN, and 2) region-based attention with feed-forward networks for refining attentional features.  Following Zhong et al. \cite{zhong2020random},  random erasing  image augmentation has been adapted here on-the-fly to mitigate  overfitting. The proposed method is briefed in Algorithm \ref{alg:Algo1}, wherein steps 1-4 represent hybrid augmentation and steps 5-17 define successive operations of RAFA-Net. The specific step numbers are mentioned in stage-wise description below.

\subsection{Stage 1: Region Based Deep Feature Representation}
This stage involves random erasing data augmentation and feature extraction using a backbone CNN, and computing local regions via pooling, those are described next.

\subsubsection{Hybrid Random Erasing Data Augmentation}
This technique is applied for generating additional training data that could be tailored to different model capacities and dataset sizes. Random erasing is postulated  by removing a region that is chosen stochastically with height $h$ and width $w$  within the input image $I$, and replacing the actual pixel values with some random values with a certain probability \cite{zhong2020random}.  A random function $Rnd$() is used to select the height $\Delta_h$ and width $\Delta_w$ of a region, which is to be erased with a pixel value $q$ to distort a mini-batch. An initial pixel $(s_x, s_y)$ is chosen randomly with respect to which a sub-region is wiped off such that an occluded region should contained within  $I$. 
Moreover, random rotation and cropping are applied  for additional distortions upon stochastic erasing. This hybrid scheme leveraging both types of augmentation enforces better generalization ability of the network,  albeit incurring no extra computation cost during inference. The hybrid strategy augments a local region followed by an explicit global transformations that helps to tackle the  sub-optimality of either augmentation method. 
\RestyleAlgo{ruled}
\begin{algorithm} 
\caption{Proposed RAFA-Net } \label{alg:Algo1}
\textbf{Input:} {image $I$, RoIs $R$, learning parameters $\theta$} \\
\textbf{Output:} {predicted class-label $Z_{pred}$ of objects } \\

1. $\Delta_h\gets Rnd(0.2, 0.7)\times h$; $\Delta_w\gets Rnd(0.2, 0.7)\times w$;\\
2. $s_x \gets Rnd(0, h)$; $s_y \gets Rnd(0, w)$; such that\\
 \hspace{0.3 cm} $s_x+\Delta_h \leq h$, and  $s_y+\Delta_w \leq w$\;
3. $I [{s_x, s_y, s_x+\Delta_h, s_y+\Delta_w]} \gets q$; /* erased val=\textit{q} */ \\
4. $I^*\gets Augment(I)$;  /* global image augment */ \\
5. \textbf{Initialize} CNN with the parameters $\theta$

\Repeat{$\theta$ converges}{
6. ${F}\gets BaseDeepNetwork(I^*)$ \\
7. ${F_u}\gets Upsampling(F)$ \\
8. ${F_R}\gets PoolingRoI({F_u}, R)$\\
9. ${F_{LN}}\gets LayerNorm(Attention(F_R))$  \\
10. ${F_{1}}\gets SepConv_{1D}(AvgPool(F_{LN}))$ \\
11. ${F_{2}}\gets SepConv_{1D}(PyramidPool(F_{LN}))$ \\
12. ${F_{sum}}\gets AvgPool(F_1)+AvgPool(F_2)$\\
13. ${F_A}\gets LayerNorm({F_{sum}})$ \\
14. ${F_{M}} \gets F_{A} +F_{A}$$ \otimes  \big(\sigma(AttentionWgt(F_{1}))\big )$ \\
15.  ${F_{final}}\gets LayerNorm(Dropout({F_{M}}))$\\
16. ${Z_{pred}}\gets Softmax(LayerNorm({F_{final}}))$\\
17. compute loss $\mathcal{L}_{ce}$ and optimize parameters $\theta$;
}
\textbf{end}
\end{algorithm}

\subsubsection{Contextual Regions Formulation} 
Contextual information provides a vital cue for learning  the semantics from various regions of an image.  In CNN architectures of some existing methods,  the relationship between different local features have been ignored which are relevant for classification \cite{sheng2024lightweight}, \cite{wu2024strawberry}. Here, the aim is to  interpret finer details of underlying patterns from complementary regions to define a holistic feature map. A uniform  representation of regions explores and enriches subtle details, which are  enhanced further by paying attention to relevant activations in the feature space. The formulation of  regions depends on the spatial dimension of an output feature vector computed from a  base CNN.

\vspace{-0.1 cm}
Let an input visual image $I_z$ $\in$ $\mathbb{R}^{h\times w\times 3}$ with its class label $z$ is fed into a base CNN for extracting a  high-level convolutional feature map $\textbf{F}^l$ $\in$ $\mathbb{R}^{h\times w\times c}$ at layer $l$ where $h$, $w$, and $c$ refer to the height, width, and channels, respectively (steps 5-6).  A base CNN reduces the spatial dimension of an input image drastically, from which extraction of region-level locality with a standard resolution is difficult. To fix this issue,  upsampling the  spatiality of a feature map using interpolation is  applied. The upscaled dimension is adjustable  as required to define  structural neighborhood by  distinct fixed-size regions. $\textbf{F}^l$ is divided into smaller regions for learning the details of local representation. The size of a region infers a window size for region-based pooling of   $\textbf{F}^l$.      
A region is defined as $d_r=[x_r, y_r, \delta w, \delta h]$,  where $(x_r, y_r)$ denotes an initial pixel of a region; and the window-size $\delta w=\delta h=\delta$ with $\delta^2$ area is the same for all regions. 
The number of  regions is given as $R={(h\times w)}/\delta^2$. 
The feature maps of  $R$  at layer $l$ is denoted as $\textbf{F}^l=\big\{\textbf{F}_r\big\}_{r=1}^{r=R}$ $\in$ $\mathbb{R}^{R\times (h\times w\times c)}$, 
computed by relating a region with the feature map. In this regard, $\textbf{F}^l$ is scaled up to a higher dimension. A bilinear interpolation is used by means of a mapping: ${\textbf{F}} \rightarrow {\textbf{F}}\in \mathbb{R}^{k(h\times w)\times c}$, where actual resolution $(h$$\times$$w)$ is upsampled by $k$ times, and then, each region is extracted accordingly (steps 7-8). The feature dimension after region pooling is similar to base CNN  \textit{i.e.}, $\textbf{F}^l\in$ $\mathbb{R}^{(h\times w\times c)}$.  
 
In this way, stage-1 strengthens  input-data variability using augmentation prior to compute a high level feature map  using a base CNN. Then, the global feature summary is endowed with local description by extracting regional information.

\subsection{Stage 2: Attention Based Feature Refinement via FFNs }\label{sec:attn}
This stage  describes the main components of the proposed model, consisting of a region-based attention method, feature transformations through dual FNNs, and  feature refinement with attention weights. To this end,  these modules produce a refined feature map that is used for classification.

\subsubsection{Region Attention}\label{sec:Regattn}
Attentional feature description represents   semantic understanding by cross-channel interactions and builds a strong  correlation among contextual regions. A channel-wise relationship  attends   crucial regions for enhancing feature aggregation ignoring lesser significant ones, and thereby improves  image recognition performance.

Self-attention  \cite{vaswani2017attention}  solves long-distant dependency through  a context vector computed according to a weighted sum of feature maps. It allows content-based interactions and  parameter-independent scaling of receptive field size which is suitable for solving vision-related challenges \cite{vaswani2021scaling}. Here, self-attention is applied across the channels for regional attention  due to its superior performance over others (step 9). 

Self-attention uses three similar feature maps, $[\textbf{Q}, \textbf{K}, \textbf{V}]$, the {query} \textbf{Q}, {key} \textbf{K}, and {value} \textbf{V}  which are learned from the same feature map $\textbf{F}^l$. 
The attentional weight matrix is derived by  a dot product of \textbf{Q} and \textbf{K}, which is multiplied with \textbf{V}.  
Here, $[\textbf{Q}, \textbf{K}, \textbf{V}]$ vectors are computed for region attention. The attentional importance is determined using  feature vectors $\textbf{F}_i$ indicating the  $i^{th}$ region and $\textbf{F}_j$ implying a neighboring $j^{th}$ region ($i\neq j \in R$). 
The attention score indicates the relevance of a region stipulated on other candidate regions,  defined as
\begin{equation}
\centering
\begin{split}
    \mathbf{G}_{i,j} &={tanh}({\mathbf{W}_G \textbf{F}_i} + {\mathbf{W}_{G'}\textbf{F}_{j}}+\mathbf{b}_{G}), \\
        \mathbf{H}_{i,j}&=sigmoid\left(\mathbf{W}_H \mathbf{G}_{i,j} +  \mathbf{b}_{H} \right) \\
\end{split}
\end{equation}
where $\textbf{W}_{G}$ and $\textbf{W}_{G'}$ are the weight matrices for computing attention vectors using $i^{th}$ and $j^{th}$ regions, respectively;  $\textbf{W}_{H}$ denotes a nonlinear combination; $\mathbf{b}_{G}$ and $\mathbf{b}_{H}$ imply the biases; and $sigmoid(.)$ is an element-wise  activation function for  nonlinearity. The importance of each region is computed by a weighted sum of  attention values considering all regions. 
\begin{equation} \label{eqx2}
\centering
\begin{split} 
M_{i,j}={softmax}(\mathbf{W}_{M}H_{i,j}+\mathbf{b}_M), \text{ }
\tilde{\mathbf{F}}^l_i=\sum_{j=1}^{{R}}&M_{i,j}\textbf{F}_{j} \\
\hat{\mathbf{F}}_i^{(l+1)}= {LayerNorm}(\tilde{\mathbf{F}}^{l}_i; \alpha_i, \beta_i)
\end{split}
\end{equation}
where $\textbf{W}_{M}$ denotes the weight matrix, and the bias is ${b}_M$. The feature map at layer $l$ is $\hat{\textbf{F}}_i^l$  that has been passed through a layer normalization (LayerNorm) at layer $(l+1)$, denoted  as $\hat{\textbf{F}}_i^{l+1}$,    $\forall r_i \in R$.
LayerNorm is a better alternative to batch normalization, as the latter depends on a mini-batch size \cite{ba2016layer}. LayerNorm maintains the invariance properties (\textit{e.g.}, scaling, shifting of weight matrix) within a layer during training and fixes the mean and variance of the summed inputs within each layer.  As a result, LayerNorm reduces the “covariate shift” problem, and improves training speed. In general, LayerNorm is defined as a mapping function with two sets of adaptive parameters, gains $\alpha$ and biases $\beta$. It is defined as
\begin{equation}
\begin{split}
{LayerNorm}(\mathbf{F}^l;\alpha, \beta)= \frac{(\mathbf{F}^l-\mu^l)}{\sigma^l}\odot \alpha +\beta \text{,   where  } \\
\mu^{l} =\frac{1}{L}\sum_{k=1}^{L}\textbf{F}_{k}^l, \text{  and    } \sigma^{l}=\sqrt{\frac{1}{L}\sum_{k=1}^{L} (\textbf{F}_{k}^l-\mu^l)}  
\end{split}
\end{equation}
where ${\textbf{F}}^l$ implies the feature map at layer $l$, and $L$ signifies the number of hidden units in a layer.  The mean  $\mu$  and standard deviation $\sigma$ both indicate normalization terms of LayerNorm. Here,  LayerNorm is used  for robustness over mini-batch sizes and faster training convergence. 
\subsubsection{Feed-Forward Network (FFN)} 
It is a vital component of a deep  model for feature enhancement. A few existing works used a multi-layer perceptron (MLP) as a FFN. A MLP depends on an attention mechanism  for computing spatial relations (\textit{e.g.}, inter-pixel dependencies) in the feature space \cite{wu2022p2t}. Self-attention is suitable for correlating long-distant content globally, which sometimes, fails to 
capture local content interaction. In general, this type of FFN  is not enough for learning  spatial locality from  complex contexts, such as  food patterns and layouts of different plant lesions.
This problem can be tackled locally by a convolution with a modest receptive field size. The  local inductive bias and translational equivariance nature of standard convolution could overcome this issue. Thus,  a depth-wise separable convolutional (SepConv) layer  is  injected in FFN that integrates the benefits of attention  for long-range dependency modeling; and  SepConv for correlating channel-wise and spatial interactions.  

In  depth-wise convolution,  spatial convolution is applied to each channel independently, followed by  point-wise convolution transforming the channels into a new feature map  \cite{chollet2017xception}. This technique for decoupling  a convolutional feature map in two different dimensions essentially improves learning  performance. Yet, SepConv enjoys a low-rank factorization for spatial and channel-wise interactions.   
Here, a 1D SepConv layer is included with $relu$ activation and \textit{kernel-size}=3.  It is expected  that in a deep network  highly correlated spatial structures could be captured from local neighborhood information with a modest receptive field size without much information loss during aggregation via non-linear activations. However, increasing excessive activations within a FFN may cause substantial disentangled features, which is avoided by a single 1D SepConv layer in  FFN. This layer is applied after pooling attention features in two  paths to integrate the benefits both for feature selection (steps 10-11). 

\noindent \textit{Spatial Pyramid Pooling and Global Average Pooling}:
Spatial pyramid pooling (SPP) describes local structural information of a feature map in a hierarchical fashion by varying filter-sizes  \cite{he2015spatial}. It  allows variable image dimensions and aspect ratios. Here,  window sizes of  [$1\times1$; $2\times2$; and $3\times3$] representing a total of 14 bins are selected in a pyramid hierarchy for pooling from attentional features, and produces output vector $\hat{\mathbf{F}}_{A}$. 
\begin{equation}
\centering
\begin{split}
\hat{\mathbf{F}}_{Ai}^{(l+1)}= {PyramidPool}(\hat{\mathbf{F}}_i^l) \text{  ;  } \hat{\mathbf{F}}_{Bi}^{(l+1)}= {AvgPool}(\hat{\mathbf{F}}_i^l) 
\end{split}
\end{equation}
A global average pooling (GAP) layer that performs complementary of SPP is added in another path. The GAP explicitly considers full feature map as a whole, and  summarizes overall attentional feature maps of multiple regions. Here, a GAP layer is useful, albeit  SPP does not consider a (1$\times$1) bin. It implies that even with a single pyramid level (\textit{e.g.}, $3\times3$), which is useful for accelerating the training phase, GAP induces the whole feature map into the landscape complementarily for aggregation. The GAP generates a refined descriptor $\hat{\mathbf{F}}_{B}$ which improves descriptiveness by summarizing  semantic information from each region. Overall, it offers a global feature aggregation at different contexts from multiple regions. 

The pooled feature maps are computed at two different aspects, \textit{i.e.}, one from considering all smaller regions uniformly; and another considering pyramid-levels for focusing local information. Both pooling techniques are combined for representing a powerful feature descriptor. To this end, two similar FFNs are developed for capturing  and mixing channel and spatial correlation by SepConv, as described earlier. 
\begin{equation}
\centering
\begin{split}
{\mathbf{S}}_{R}^{(l+1)}= {SepConv}(\hat{\mathbf{F}}_{A}^l) \text{  ;  } {\mathbf{T}}_{R}^{(l+1)}= {SepConv}(\hat{\mathbf{F}}_{B}^l) 
\end{split}
\end{equation}
\begin{equation}
\centering
\begin{split}
\hat{\mathbf{S}}^{(l+1)}= {AvgPool}({\mathbf{S}}_{R}^l) \text{  ;  } \hat{\mathbf{T}}^{(l+1)}= {AvgPool}({\mathbf{T}}_{R}^l) 
\end{split}
\end{equation}
\begin{equation}
\centering
\begin{split}
{\mathbf{X}}^{(l+1)}= {LayerNorm}({Add}(\hat{\mathbf{S}}^l \text{  ;  } \hat{\mathbf{T}}^l) ) 
\end{split}
\end{equation}

The convoluted feature maps of both paths are reshaped and passed through similar 1D GAP  layers to summarize  features across the channels. Then, the outcomes of both paths are added to conflate their benefits via  LayerNorm (steps 12-13).

\subsubsection{Attention-Weighted Feature Refinement \& Classification}
Next aim is to refine channel-wise feature interactions by attention-based weighted score. For this intent, ${{\mathbf{T}}}_R$ is transformed by a $softmax$ activation to estimate a weighted-attention matrix $\Phi_{i}$. Then, a weighted sum  $\tilde{\mathbf{T}}_{A}$ is computed as an outcome of inter-channel attention. A similar description of weighting the attended features is  defined earlier (Eq.\ref{eqx2}).
\begin{equation}
\centering
\begin{split}
{\tilde{\mathbf{T}}_{A}}=\sum_{i=1}^{{R}}&\Phi_{i}{{\mathbf{T}}}_i 
\text{ where,} \hspace{1 mm} \Phi_{i}={{softmax}}(\mathbf{W}_{\Phi}{{	{\mathbf{T}}}_i}+\mathbf{b}_\Phi) 
\end{split}
\end{equation}
An identity mapping with fused pooled features is included  at a deeper layer (before classification) to ease training degradation problem and  emphasize weighted features.  Here, the  layers are added relying on an intuition that  a deeper model should not have greater training error compared to its shallower counter design. As a  plausible remedy,  an element-wise addition with non-linearity is added as a residual path to alleviate training degradation  and improve learning efficiency  (step 14).

A simple context gating  is applied for weight re-calibration that incorporates non-linearity by an element-wise $sigmoid$ activation and multiplication. This gating preserves meaningful attentional features by re-weighting their relevance for final refinement and suppressing triviality before a $softmax$ layer. 
\vspace{-0.2 cm}
\begin{equation}
\centering
\begin{split}
{\mathbf{X}_{A}}={LayerNorm}{\bigl( {sigmoid}(\tilde{\mathbf{T}}_A) \otimes \mathbf{X} +\mathbf{X} \bigr)}\\
{\mathbf{Z}_{pred}}={softmax}{\bigl(\varphi(\mathbf{X}_A) \bigr)}
\end{split}
\end{equation}

A regularization layer is added to generalize  model capacity (step-15). A Gaussian dropout (GD) layer, denoted as $\varphi$,  regularizes the  model by inducing an upper bound on the mutual information between the input and output layers \cite{reygaussian}. It uses multiplicative Gaussian noise as a better alternative than traditional dropout layer to mitigate overfitting without additional overhead,  \textit{i.e.}, free from any learnable parameter. This hyper-parameter is specified  as the noise standard deviation $\varphi_{noise}(q)=\sqrt{q.(1- q)^{-1}}$, where  $q$ is a a drop-out rate between [0, 1]. 
The  error rate between a correct class-label ($Z_{true}$) and predicted class-label ($Z_{pred}$) is reduced  during model training by categorical cross-entropy loss function $\mathcal{L}_{ce}(Z_{true}, Z_{pred})$,  given in steps 16-17 of Algorithm \ref{alg:Algo1}.

In summary, the proposed region attention exploring multi-scale pooling enhances overall feature representation, which is further guided by weights to render a robust description of input samples of multiple domains. In this manner, agro-food classification tasks which are brought under one umbrella have been tackled with our  RAFA-Net.

\vspace{-0.2 cm}
\section{Experimental Results and Discussion} \label{experiments}
A summary of food datasets used for experiments, followed by the implementation specification, overall performance evaluation, and comparative studies are  presented here. Dataset summary with the best top-1 accuracy (\%) attained by RAFA-Net  is showcased for a comparison with state-of-the-arts (SOTA) in Table \ref{exp1}. The DenseNet-169 is the best performer on the UECFood-100 dataset, and Xception has achieved the highest accuracy  for the rest all datasets. 

\begin{figure*}
\centering
\includegraphics[width=0.25\textwidth, height=2.8 cm]{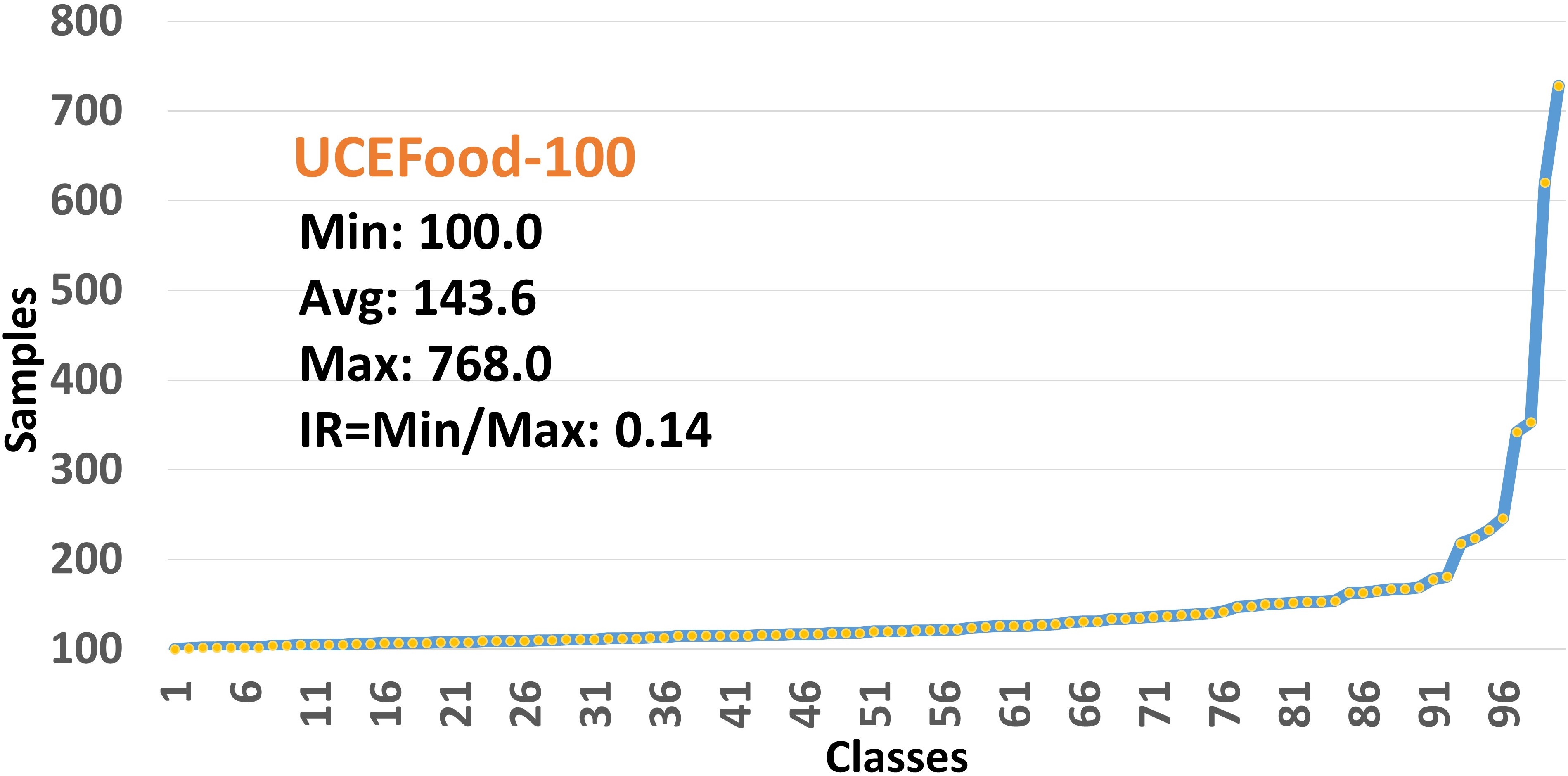} \hfill
\includegraphics[width=0.25\textwidth,  height=2.8 cm]{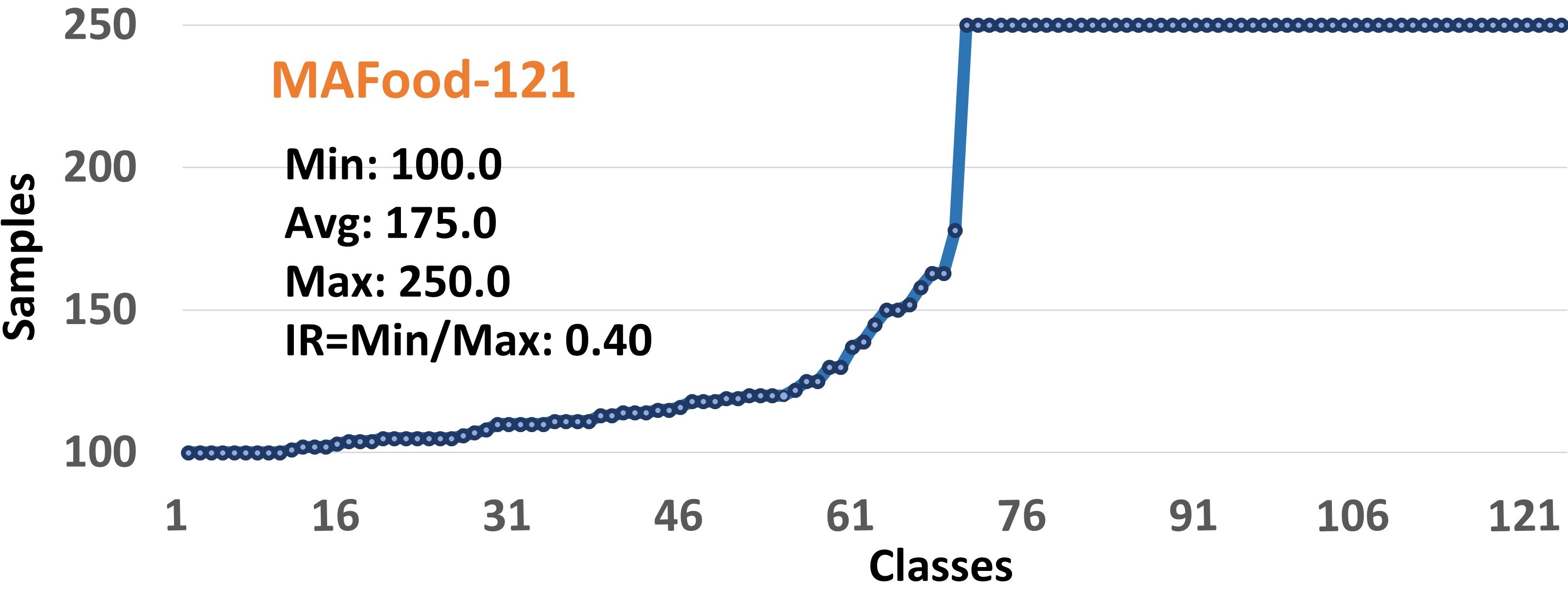} \hfill
\includegraphics[width=0.22\textwidth,  height=2.8 cm]{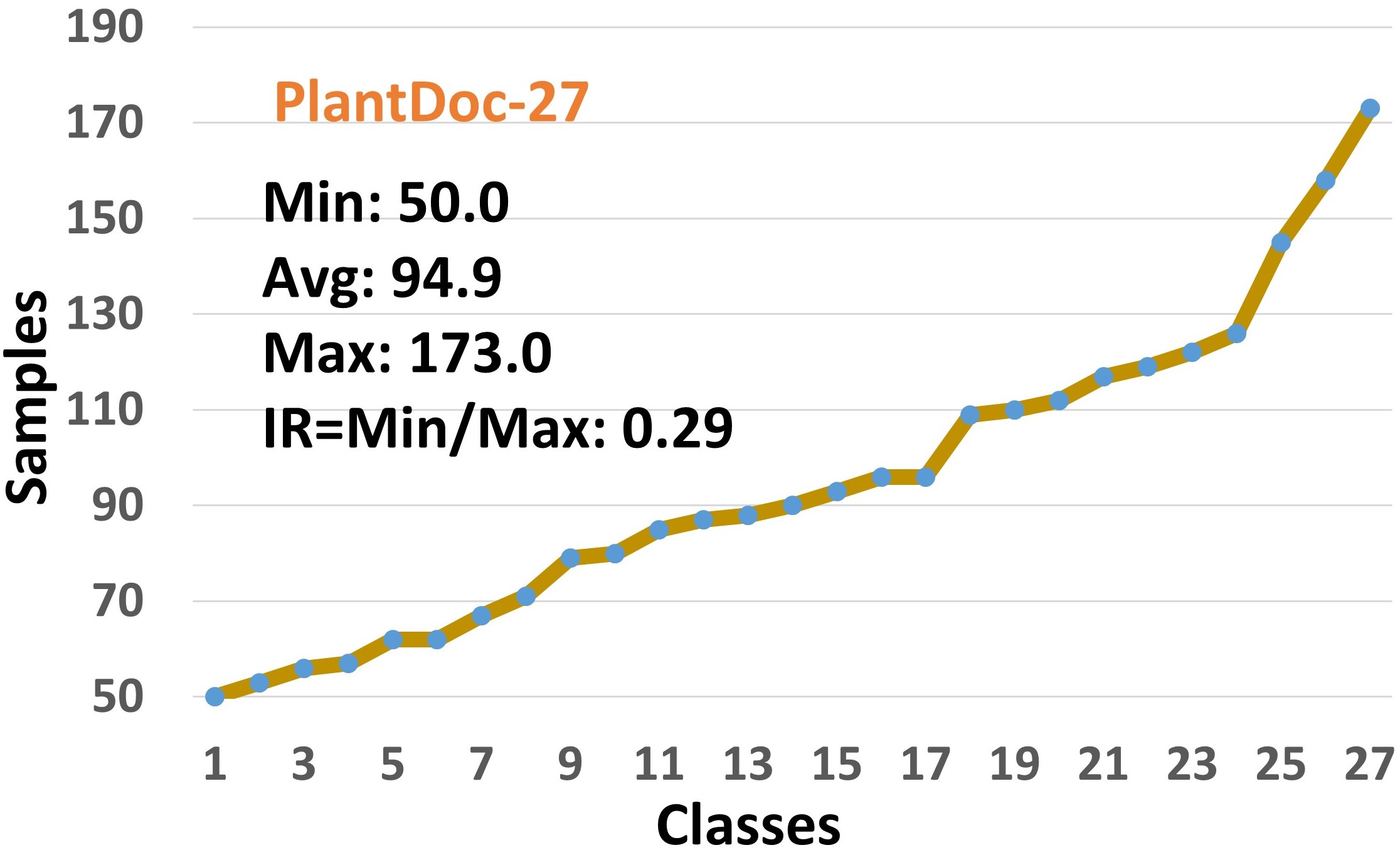}\hfill
\includegraphics[width=0.24\textwidth,  height=2.8 cm]{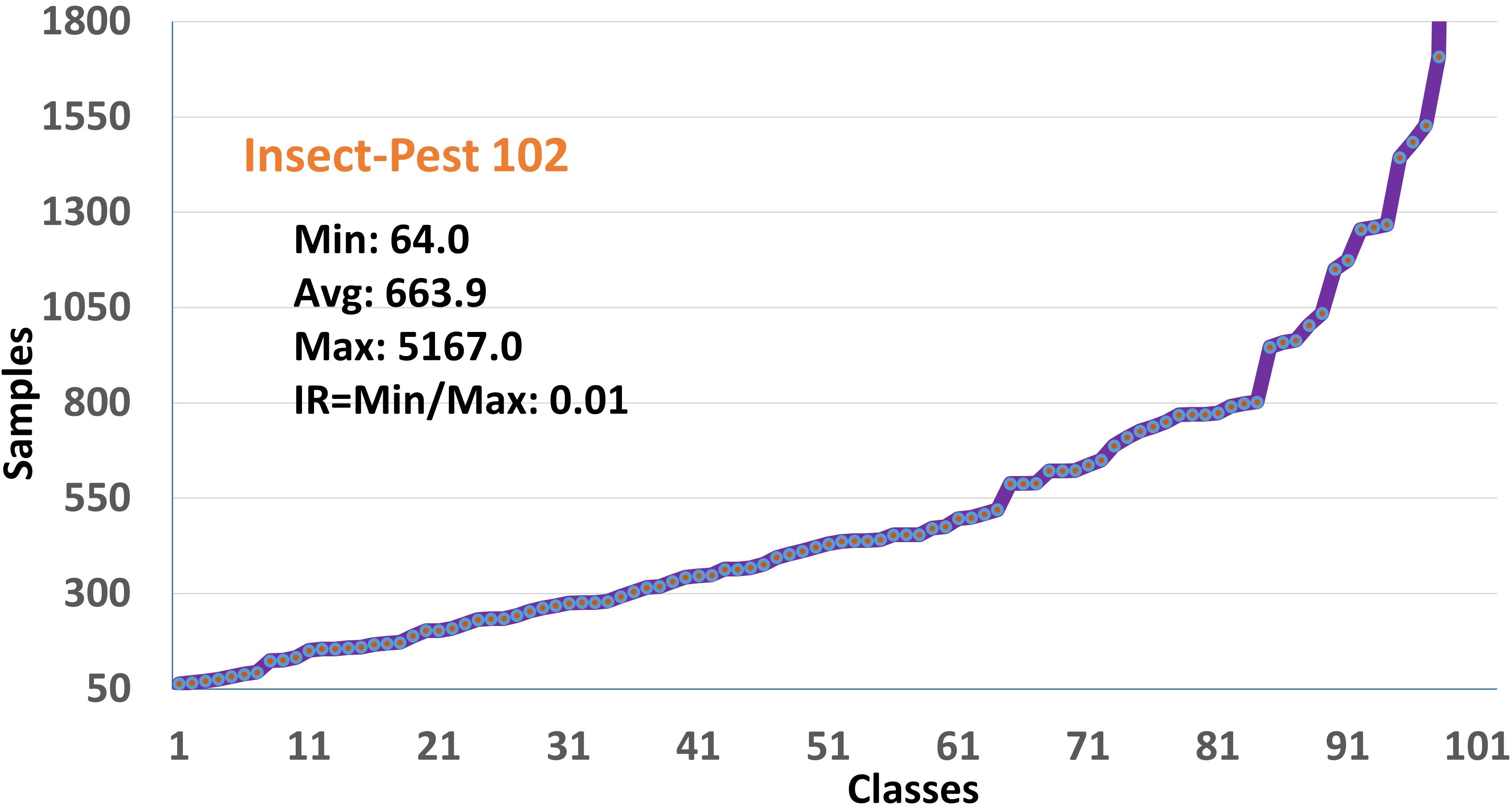} \hfill
\caption{Class-wise image distributions of the  UCEFood100, MAFood121,  PlantDoc27, and IP102 datasets. The curve of UCEFood256 is akin to UCEFood100.}  
\label{DDM121}
\end{figure*}

\subsection{Dataset Summary} 
The RAFA-Net is evaluated on five  benchmark datasets.  A brief  characteristics of the datasets including the sample sizes of train-test split, and class imbalance ratio (CIR)  are  summarized  in Table \ref{exp1}. The CIR is determined  as the ratio of sample size of the smallest minority  class and that of the  largest majority class. CIR lies between  [0, 1], a high CIR value implies less imbalanced classes. The class-wise sample distributions of each datasets is shown in Fig. \ref{DDM121}. To address the influence of CIR in performance, commonly used metrics are the precision, recall, and F1-scores, defined in \cite{wu2024strawberry}. These metrics are also used in our work for better predictive analysis. 

\begin{figure}
\centering
\subfloat[UECFood-100]{
\includegraphics[width=0.16\textwidth, height=2.3 cm]{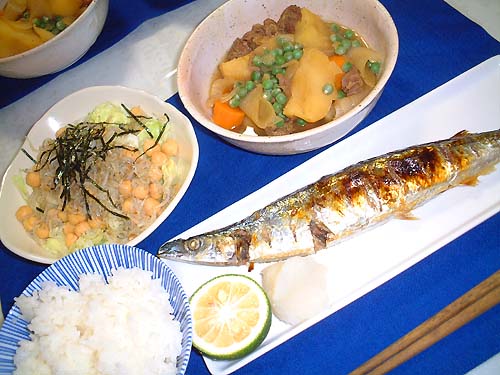} \hfill
} %
\subfloat[UECFood-256]{
\includegraphics[width=0.16\textwidth, height=2.3 cm]{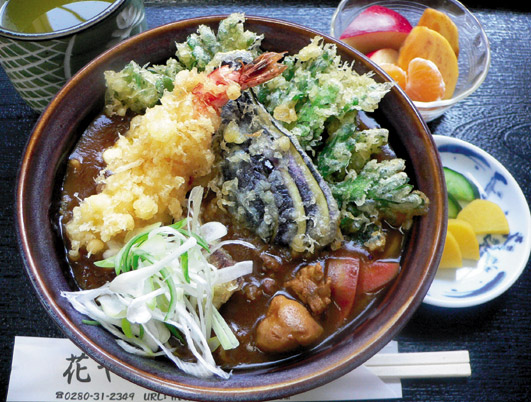} \hfill
} %
\subfloat[MAFood-121]{%
\includegraphics[width=0.16\textwidth, height=2.3 cm]{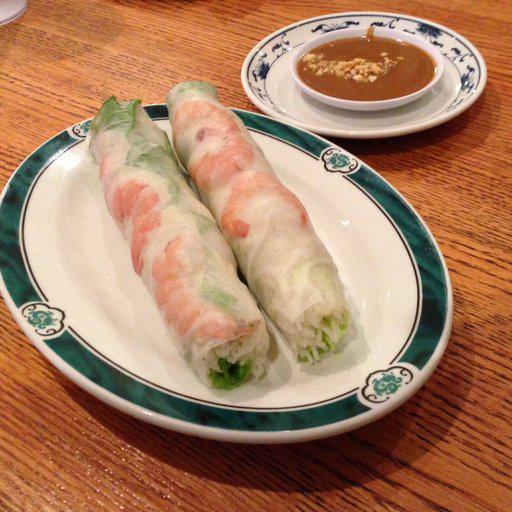}}\hfill
\caption{Sample dishes from the UECF-100\& 256, and MAF-121 datasets}
\label{UECF100}
\vspace{-0.2 cm}
\end{figure}

\noindent\textbf{UECFood-100} \cite{matsuda2012recognition}:  
It represents a Japanese food dataset with 14k images of 100 different dishes. Sample food dishes are shown in Fig. \ref{UECF100}a. The CIR=0.14, and the mean sample size is 143.6, and only 8 classes have more than 200 samples each.  

\noindent\textbf{UECFood-256} \cite{kawano2015automatic}: 
It comprises 31.5k images of 256 food categories of different countries (Fig. \ref{UECF100}b). Food dishes from different cuisines \textit{e.g.}, the Chinese, Japanese, Western, etc. are included.  The mean sample size is 123.6, containing the same 8 classes, each having more than 200 images like UCEF100, implying similar image distribution and CIR=0.14. 

\noindent\textbf{MAFood-121} \cite{aguilar2019regularized}: This  Multi-Attribute Food dataset is a recent public  dataset  categorized into 121 classes (Fig. \ref{UECF100}c). It  consists  of 21.1k images with annotations. It has a decent CIR=0.40 with the mean sample size is 175, and 43.8\% classes have the maximum 250 samples.

\noindent \textbf{IP-102} \cite{wu2019ip102}: A large benchmark dataset containing 102  classes of insect pests with 75.2k samples, shown in Fig. \ref{fig:DB1a}c. 
IP-102 consists of imbalanced class  distribution due to hierarchical representation and variations in class-wise samples. The CIR= is 0.01 that causes due to only 15\% classes have more than 1000 samples, whereas the mean  sample size is 664. 

\noindent \textbf{PlantDoc-27}  \cite{singh2020plantdoc}: 
A complex agricultural dataset curated in uncontrolled environments,  comprising of 27 classes with 2.6k original samples of plant diseases is tested (Fig. \ref{fig:DB1a}b). It has an  average class sample size of 95 and a reasonable CIR=0.29.  

\begin{table}
\centering
\caption{{Dataset Summary and the Best Top-1 Accuracy of the Proposed RAFA-Net. CIR Implies Class Imbalance Ratio within [0, 1]}  } \label{exp1}
\begin{tabular}{c c c c c c c}
\hline
Dataset & Class & Train & Test & {CIR} &SOTA  & Proposed \\
\hline
UECF100 & 100 & 11504&2857 & {0.14} &  90.02  \cite{arslan2021fine} & \textbf{91.69}  \\
UECF256 & 256 &  24974& 6421 &{0.14}  & 84.00 \cite{zhao2020jdnet} &  {\textbf{91.56}}    \\
MAF121 & 121   &15298  & 5877 &{0.40} & 88.95  \cite{arslan2021fine} &  \textbf{96.97}  \\
\hline
IP102 & 102   & 45095& 22619 & { 0.01} & 74.90  \cite{peng2022cnn} &  \textbf{92.36}  \\
PlantDoc & 27   & 2047& 516 & {0.29} & 84.56 \cite{pal2024robust} &  \textbf{85.54}  \\
\hline
\end{tabular} 
\vspace{-0.2 cm}
\end{table}

\begin{table}
\centering
\caption{{Performance on the UECFood-100  using Various Metrics.}  }\label{tab:UECF100metrics}
\begin{tabular}{c c c c c  c} 
 \hline
 Base CNNs & Top-1  & Top-5 & Precision   & Recall & F1-score \\           \hline
MobileNet-v2  & 91.27 & 98.14& 92.0 &92.0 & 92.0\\
Inception-v3  &91.24 &98.42 &91.0  &91.0 &91.0 \\
ResNet-50  & 91.28 & 98.56 & 91.0 & 91.0 &91.0\\ 
DenseNet-169  & \textbf{91.69} &98.63 & 92.0& 92.0&92.0\\ 
Xception  & 91.10 & 98.24  & 92.0 & 91.0 & 91.5  \\    
\hline
\end{tabular}
\vspace{-0.1 cm}
\end{table}

\begin{table}[!htp]
\centering
\caption{{Performance on the UECFood-256 using Various Metrics.}  }\label{tab:UECF256metrics}
\begin{tabular}{c c  c c c  c c} 
 \hline
 Base CNNs & Top-1  & Top-5 & Prec.   & Recall & F1-scr & Par (M) \\           \hline
MobileNet-v2 & 89.80 & 97.96 & 90.0 & 90.0 & 90.0 & 10.0 \\
Inception-v3 &90.38 &98.06  & 91.0 &90.0 &  90.5 &37.3\\
ResNet-50  & {90.47} & 98.34 & 91.0 & 91.0 &91.0  & 39.1 \\ 
DenseNet-169  & 90.29  & 97.06 & 91.0 & 90.0 &90.5 & 24.0\\ 
Xception   & \textbf{91.56} & 98.86  & 92.0 &92.0  & 92.0  & 36.4 \\    
\hline
\end{tabular}
\vspace{-0.1 cm}
\end{table}

\begin{table}[!htp]
\centering
\caption{{Performance on the MAFood-121 using Various Metrics.}  }\label{tab:MAF121metrics}
\begin{tabular}{c c  c c c  c} 
 \hline
 Base CNNs  & Top-1  & Top-5 & Precision   & Recall & F1-score \\           \hline
MobileNet-v2 & 96.38 &99.21  & 96.0 & 96.0 &96.0 \\
Inception-v3  &96.88 &99.42  & 97.0 & 97.0  &97.0 \\
ResNet-50 & 96.91 & 99.62 & 97.0 & 97.0 &97.0  \\ 
DenseNet-169  & 96.61  &  99.33 & 97.0 & 97.0  &97.0 \\ 
Xception   & \textbf{96.97} &  99.64 & 97.0 & 97.0  &97.0 \\    
\hline
\end{tabular}
\vspace{-0.2 cm}
\end{table}

\subsection {Implementation Specification} \label{implemnt}
Pre-trained \textit{ImageNet} weights are used for initializing base CNNs \textit{e.g.}, DenseNet-169, ResNet-50, etc. An input image is provided with class label having resolution 256$\times$256.  An image region with size (0.5$\pm$0.2) is wiped off with RGB value 127. Common augmentation methods \textit{i.e.},  random rotation ($\pm$25 degrees) and scaling ($\pm$0.25)  are applied, and  cropped with image size of 224$\times$224. The output feature map of a base CNN is spatially resized to 48$\times$48 before region pooling. The size of each region is set to 16$\times$16 pixels for $R=9$. Total 14 bins are generated by SPP with sizes $[1\times1, 2\times2, 3\times3]$. A Gaussian dropout rate of $0.25$ is set to ease overfitting. The optimization of categorical cross-entropy loss is tackled by the Stochastic Gradient Descent (SGD) optimizer. The  RAFA-Net  is trained for 100 epochs with a starting  learning rate of $0.008$,  and then  divided by 10 after 50 epochs. The training is performed using a  8 GB Tesla M10 GPU with a mini-batch size of 8. The model parameters are calculated in millions (M).

\subsection{Food Recognition Result Analysis and Discussion}
The efficiency of RAFA-Net is evaluated using five base CNNs which are built with intrinsically diverse architectural design families having a wider range of parametric complexities, such as residual connections (ResNet-50), densely connected  (DenseNet-169), separable convolution (Xception), inception modules (Inception-v3), and  
lightweight MobileNet-v2. The  top-1 and top-5 accuracy, precision, recall, and F1-score are computed and  given in Tables \ref{tab:UECF100metrics}-\ref{tab:MAF121metrics}.

The best 91.69\% top-1 accuracy on the UECF100 dataset is attained by DenseNet-169. Other backbones have obtained very competitive and decent performances. The results are detailed in Table \ref{tab:UECF100metrics}. Whereas, Xception performs the best in recognizing  food categories of UECF256 which is a larger dataset compared to the UECF100 and MAF121. The highest accuracy attained by Xception  on the  UECF256 is 91.56\%, and overall results are reported in Table \ref{tab:UECF256metrics}. The performances on the MAF121  dataset are given in Table \ref{tab:MAF121metrics}, implying the highest accuracy is 96.97\% attained by Xception. Like other datasets, the performance variations of using different CNNs are marginal, and all backbones have  attained improved top-5 accuracy on these food datasets. 

\begin{table*}
\caption{Comparison of Top-1 and Top-5 Accuracy  on the UECFood-100,  UECFood-256  and MAFood-121 Datasets } \label{Comparson_all}
\begin{center}
\vspace{-0.2cm}
\begin{tabular}{ ccc|ccc |ccc } 
\hline
 \multicolumn{3}{c}{UECFood-100} & \multicolumn{3}{|c|}{UECFood-256} & \multicolumn{3}{c}{MAFood-121} \\
 Method &  Top-1 & Top-5  & Method &  Top-1  & Top-5  & Method &  Top-1  & Top-5 \\
\hline
DCNN-Food (ft2) \cite{yanai2015food} & 78.77 & 95.15  & DCNN-Food (ft) \cite{yanai2015food} & 67.57 & 88.97 & Epistemic Uncertainty (RN50) \cite{aguilar2020uncertainty}  & 81.62 &  -   \\ 
Inception-v3 \cite{hassannejad2016food} &  81.45 & 97.27 & FS-UAMS \cite{aguilar2022uncertainty} & 73.01 & 92.58 &RUMTL+ResNet-50 \cite{aguilar2019regularized}  & 83.82 &  - \\
WARN \cite{rodriguez2019pay} &85.50 & - & Inception-v3 \cite{hassannejad2016food} & 76.17 & 92.58 & TLEnsemble (Incv3+DN201) \cite{fakhrou2021smartphone} & 84.95 & - \\
WRN \cite{zagoruyko2016wide} & 86.71 & 98.92 & WRN \cite{zagoruyko2016wide}  & 79.76 & 93.90  & ResNet-50\cite{aguilar2022uncertainty} & 83.16  &\\
WISeR \cite{martinel2018wide} & 89.58 & 99.23 & WISeR \cite{martinel2018wide} & 83.15 & 95.45 & Inception-v3 \cite{aguilar2022uncertainty} & 86.94 & -\\

Ensemble (RNX101+DN161) \cite{arslan2021fine} & 90.02& - & JDNet \cite{zhao2020jdnet}  & 84.00 &96.20   &  FS-UAMS \cite{aguilar2022uncertainty} & 88.95 &  -\\ \hline 
\textbf{RAFA-Net}   ResNet-50 & 91.28 & 98.56 & ResNet-50& 90.47 &98.34 & ResNet-50  &   96.91 &99.62 \\  

DenseNet-169 & \textbf{91.69} & 98.63 & Xception & \textbf{91.56} &98.86 & Xception   & \textbf{96.97} &99.64 \\ 
\hline
\end{tabular}
\vspace{-0.4 cm}
\end{center}
\end{table*}

\subsubsection{Performance Comparison} The WISeR  attained 89.58\% top-1 and 99.23\% top-5 accuracies on the UECF100 dataset by capturing vertical food structures using the slice convolution layer  \cite{martinel2018wide}. The best SOTA top-1 accuracy reported on this dataset is 90.02\% using an ensemble of ResNeXt-101, and DenseNet-161 \cite{arslan2021fine}. However, their method attained 86.5\% top-1 accuracy using ResNet-50. In contrast, the accuracy of RAFA-Net implemented upon ResNet-50 is 91.28\% and the best 91.69\% accuracy using DenseNet-169. Recent other works on this dataset are given in Table \ref{Comparson_all} (left). RAFA-Net outperforms  SOTA 90.02\% \cite{arslan2021fine} with  1.67\%  accuracy gain.

The JDNet attained  84.0\%  top-1 and 96.2\% top-5 accuracies on the UECF256 dataset  using a lightweight MobileNet-v2 as student and ResNet-101 as teacher networks \cite{zhao2020jdnet}.  The WISeR  achieved 83.15\%  top–1 and 95.45\% top–5 recognition accuracies on UECF256, respectively \cite{martinel2018wide}. {The GS-Net has attained its best 71.90\% top-1 accuracy  on the UECF256 \cite{sheng2024lightweight}.} A fine-tuned deep convolutional neural network  called DCNN-Food (ft)  reported 78.77\% top-1 accuracy  on the  UECF100 and 67.57\% on UECF256 \cite{yanai2015food}. It is evident that RAFA-Net has surpassed existing works on UECF256, listed  in Table \ref{Comparson_all} (middle), and achieved a 5.8\% accuracy gain using MobileNet-v2. The highest accuracy margin is  6.6\% accomplished by  the Xception-based RAFA-Net, compared to JD-Net (84.00\%).

The accuracy on the MAF121 dataset is 83.82\% using a regularized uncertainty-based multi-task learning (RUMTL) \cite{aguilar2019regularized}.  
A deep ensemble method estimating the uncertainty  has   achieved  81.62\% accuracy on the MAF121 \cite{aguilar2020uncertainty}.  The FS-UAMS \cite{aguilar2022uncertainty} reported  88.95\% accuracy using an ensemble of ResNet-50 and Inception-v3 backbones.
RAFA-Net's results on MAF121 outperform these works \textit{e.g.}, 96.91\% top-1 accuracy using ResNet-50, implying the efficiency of proposed method by exceeding 7.96\% accuracy. Moreover, the results using other metrics have been reported. Particularly, the RAFA-Net has brought off 99.62\% top-5 accuracy on the MAF121 dataset using ResNet-50 backbone.

\subsubsection{Model Design Parameters}   
The number of model parameters (M) of RAFA-Net on UECF256 are given in the last column of Table \ref{tab:UECF256metrics}. The model parameters of full RAFA-Net with 9 regions on UECF256 using MobileNet-v2 are 10.0M (lowest), and ResNet-50 is 39.1M (highest), respectively. The number of model parameters would be lesser  due to the dense layer for classifying the other two datasets, containing a lesser number of classes. For example, the RAFA-Net full model comprised with 38.8M parameters for UECF100, whereas, 39.1M parameters are required for the same  model to evaluate UECF256. The MobileNet-v2 is a lightweight model compared to others such as the ResNet, DenseNet, and other CNN families regarding model complexity. Yet, MobileNet-v2 has attained better results with a little model complexity on all food datasets. 
The model parameters  of  the ensemble  and WISeR methods are  118M  and  150M, respectively \cite{arslan2021fine}. In contrast, RAFA-Net is built with  39.1M parameters using ResNet-50, implying significantly fewer parameters  than SOTA.
\vspace{ -0.1 cm}
\begin{table}[!htp]
\centering
\caption{Top-1 Accuracy (\%) of  RAFA-Net on the Plant Stress Datasets} \label{tab:OtherDB}
\begin{tabular}{c c |c c c c } 
 \hline
Dataset & Method &ResNet & DenseNet & Xception & MobNet  \\       \hline
Insect Pest & Baseline &67.48  & 70.48 & {69.10}   &66.36\\
 & RAFA-Net&91.52  &91.21 & \textbf{92.36}   &90.31\\ \hline
PlantDoc & Baseline &64.06  &68.35 & 64.45   &64.25\\
 & RAFA-Net &  84.96 & 85.35  & \textbf{85.54}   & 84.37 \\
\hline
\end{tabular}
\vspace{-0.2 cm}
\end{table}

\subsection{Experimental Results on Plant Stress Datasets}
The results of baselines and full RAFA-Net model  on the IP102 \cite{wu2019ip102} and PlantDoc27 \cite{singh2020plantdoc} datasets are given in Table \ref{tab:OtherDB}, and samples are shown in Fig. \ref{fig:DB1a}b-c. The baseline  implies the accuracy achieved using pre-trained base CNN only, without any module of RAFA-Net (detailed in Sec.\ref{BL}).  The best top-1 accuracy of RAFA-Net on IP102 dataset is 92.36\% using Xception (baseline 69.10\%) that exceeds  SOTA 76.00\% \cite{hechen2024dilated}. Likewise, the accuracy of RAFA-Net  on PlantDoc27 is 85.15\%, and  baseline accuracy is 64.45\% using  Xception. RAFA-Net gains 4.0\% over recent SOTA 81.53\% on PlantDoc27 \cite{ahmad2023toward}. The 
overall performances of RAFA-Net on both datasets are summarized in Table \ref{tab:PDIPmetrics}, and the accuracies of RAFA-Net are higher than those of SOTA methods. 

\noindent \textit{Performance Comparison}:
The performances of existing methods on the IP102 dataset  are compared in Table \ref{tab:Insect}. The pioneering work on this dataset has reported 49.50\% using ResNet-50 and SVM classifier \cite{wu2019ip102}. A deep model built with three-way branching based on ResNet (DMF-ResNet) has attained 59.22\% accuracy \cite{liu2020deep}.  A convolutional transformer has attained 74.90\% accuracy \cite{peng2022cnn}.  The 
DWViT-ES method has achieved 76.00\% accuracy \cite{hechen2024dilated}. In contrast, RAFA-Net  outperforms  existing SOTA remarkably, with a gain of  more than 15\% accuracy on the IP102 dataset. 

\begin{table} 
\centering
\caption{{Overall Performance of RAFA-Net on Plant Stress Datasets}  }\label{tab:PDIPmetrics}
\begin{tabular}{c c c  c c c  c} 
 \hline
Dataset  & CNN & Top-1  & Top-5 & Prec.   & Recall & F1-Scr \\           \hline

Insect Pest  &  ResNet &91.52 & 99.20 & 92.0 & 92.0 &92.0 \\ 
 &  Xception & \textbf{92.36} &99.50 & 93.0& 92.0 &92.5 \\ 
 & DenseNet &  91.21 &98.93  & 92.0  &91.0 &91.5 \\

  &MobNet & 90.31  & 98.58 & 91.0  & 91.0 & 91.0 \\
\hline
PlantDoc & ResNet & 84.96 & 99.11 & 85.0 & 84.0 &84.5 \\ 

& Xception & \textbf{85.54} & 99.21 & 86.0 & 86.0 &86.0 \\ 
 & DenseNet &  85.35 &98.98  &  86.0 &86.0 & 86.0\\

  &MobNet & 84.37  & 98.82 &  85.0 & 85.0& 85.0\\

\hline
\end{tabular}
\end{table}

\begin{table} [h!]
\caption{Performance Comparison  on the Insect Pest-102 Dataset }\label{tab:Insect}
      \centering
      \begin{tabular}{c c  c} 
         \hline
      Ref   &  Method / CNN   & Accuracy (\%)  \\           \hline

Insect Pest \cite{wu2019ip102} & ResNet-50 (SVM) & 49.50\\
DMF-ResNet \cite{liu2020deep} & Multi-branch fusion & 59.22 \\
Ensemble \cite{ung2021efficient}  & ResNet50 + others  & 74.13\\
MS-ALN \cite{feng2022ms}  & ResNet-50   & 74.61\\
Fusion \cite{peng2022cnn} & CNN-ViT (ImageNet-21k)  & 74.90 \\

DWViT-ES \cite{hechen2024dilated}  & Swin
Transformer   &  76.00 \\
    \hline
    \textbf{RAFA-Net}  & ResNet-50   &{91.52} \\
            &  Xception  &\textbf{92.36} \\ 
         \hline
        \end{tabular}
        \vspace{-0.3 cm}
    \end{table}
\begin{table} [h!]
\caption{Performance Comparison on the PlantDoc-27 Dataset} \label{tab:PlantDoc}
      \centering
      \begin{tabular}{c c  c} 
         \hline
      Ref   &  Method / CNN   & Accuracy (\%)  \\           \hline

PlantDoc \cite{singh2020plantdoc} & Inception ResNet-v2 (ImageNet+PVD) & 70.53\\
 DHBP \cite{wang2022dhbp} &  SE-ResNeXt-101 (ImageNet)   & 75.06 \\
 T-CNN \cite{wang2021t}  & ResNeXt-101 (ImageNet +PVD)    & 75.58 \\
 ICVT \cite{yu2023inception} &Vision Transformer   & 77.54 \\
   Gen DL \cite{ahmad2023toward} 
 & Xception &  81.53\\
 E3B-Net \cite{pal2024robust} & Conv encoder-decoder & 84.56 \\
    \hline
    \textbf{RAFA-Net }  & ResNet-50  & {84.96} \\
            &  Xception  & \textbf{85.54} \\ 
         \hline
        \end{tabular}
\vspace{-0.2 cm}
    \end{table}
\begin{figure} [h!] 
    \centering
     \includegraphics[width=0.46\textwidth, height= 4.50 cm] {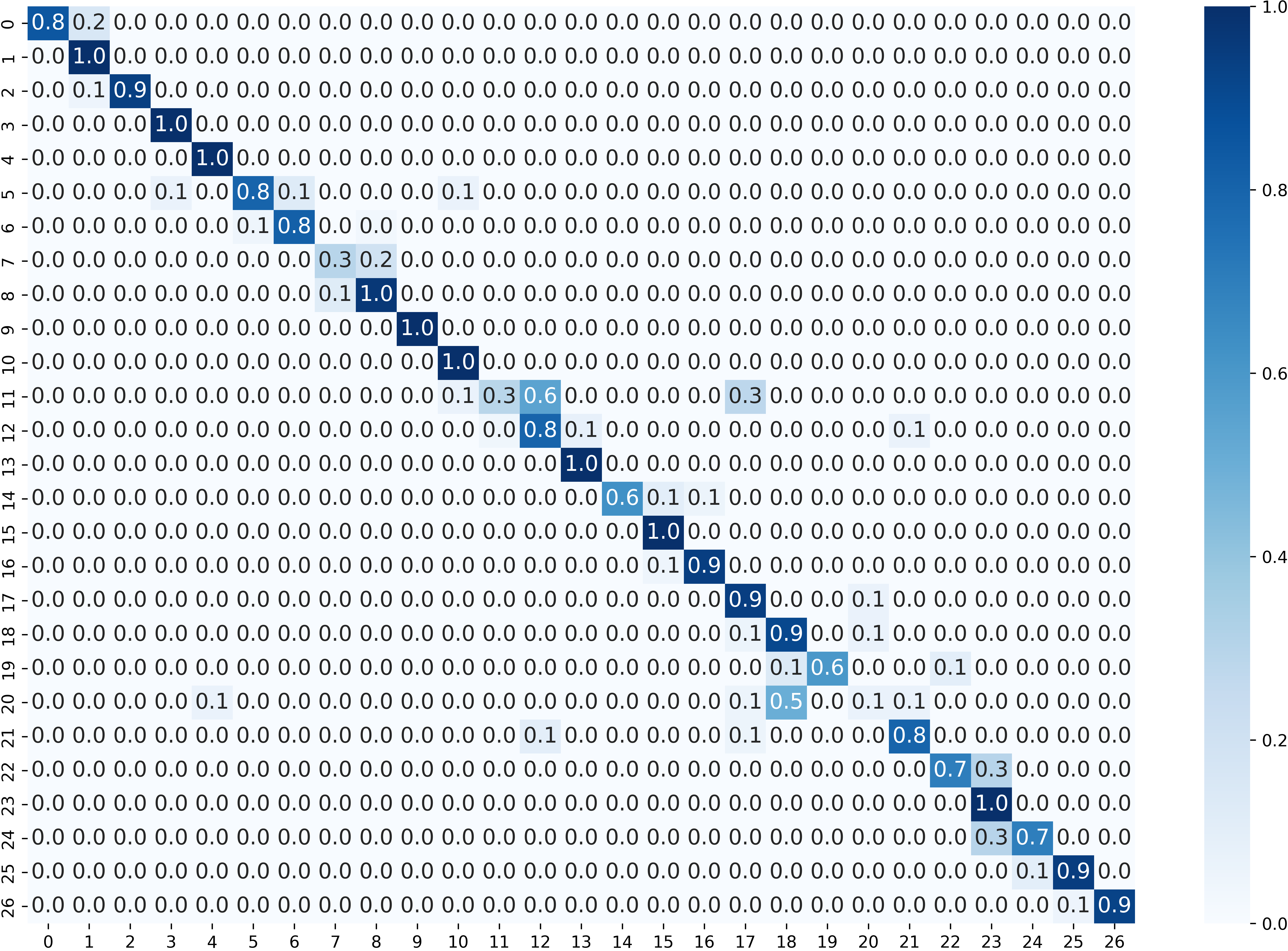}\hfill
    \caption{Confusion Matrix on PlantDoc-27 using RAFA-Net (ResNet-50) }  \label{ConMat}

\end{figure}
A comparative study on the PlantDoc27 dataset is given in Table \ref{tab:PlantDoc}. Experiments with this dataset \cite{singh2020plantdoc} has attained 70.53\% accuracy using Inception-ResNet-v2, trained with the weights of ImageNet and additional PlantVillage dataset (ImageNet+PVD). 
 A dual-stream hierarchical bilinear pooling (DHBP) using SE-ResNeXt-101 has attained 75.06\%  accuracy on this dataset \cite{wang2022dhbp}. Trilinear convolution (T-CNN) pretrained with ImageNet+PVD weights has gained 75.58\% accuracy \cite{wang2021t}.  Inception Convolutional Vision Transformer (ICVT) pretrained on ImageNet  \cite{yu2023inception} has reported 77.54\% accuracy and 77.23\% precision.  An accuracy  of 81.53\% has been attained  using Xception  on the PlantDoc \cite{ahmad2023toward}. Recently, E3B-Net has achieved 84.56\%  accuracy while tested with pretrained weights and fine-tuned on the PlantDoc27 \cite{pal2024robust}. In contrast,  RAFA-Net has brought off  85.54\% top-1  accuracy and  88.0\% precision using Xception. RAFA-Net outperforms existing methods with a significant margin  using other backbones, \textit{e.g.},  ResNet-50 has attained 84.96\% top-1 accuracy (Table \ref{tab:PDIPmetrics}). The
 training time on PlantDoc is $\approx$2.7 minutes/epoch, and 4.3 milliseconds/image during inference using MobileNet-v2. 
 A confusion matrix on  PlantDoc is shown in Fig. \ref{ConMat}. 

\begin{table}[!ht]
\centering
\caption{Top-1 Accuracy (\%) of Key Components of The RAFA-Net} \label{tab:Abln1}
\begin{tabular}{c |c c c c} 
 \hline
 Base CNNs   & UECF100  & UECF256 & MAF121  & Par(M) \\           \hline

Xception  Base  & 65.93 &  68.31 & 78.55  &21.4  \\ 
+ RoI  Attention  & 89.75 &  89.65 &  95.74 & 27.5\\ 
+  FFN (Pooling)  & 90.82 &  91.22 &  96.55  & 36.4\\ 
RAFA-Net  Full Model  & 91.10 &  91.56 & 96.97  &   36.4\\ 
\hline

ResNet-50 Base  & 63.76 &66.67 & 80.27  &  24.1\\ 
+ RoI Attention  &89.46 &88.10 & 95.35  &30.5 \\ 
+ FFN (Pooling)  &90.47 &89.37 & 96.06 &39.0\\ 
 RAFA-Net Full Model  & 91.28 & 90.47  & 96.91   &   39.1\\ 
\hline

\end{tabular}
\end{table}
\begin{table}[!htp]
\centering
\caption{Top-1 Accuracy (\%) of Feed-Forward Paths of The RAFA-Net} \label{tab:Abln2}
\begin{tabular}{c |c c c c} 
 \hline
Method (ResNet-50) & UECF100  & UECF256 & MAF121  & Par(M) \\           \hline
 $FFN_A$ (SPP) & 89.88 &88.84 &96.18 &34.9 \\
 $FFN_B$ (GAP) & 90.40 &89.46 & 96.21&34.7 \\
RAFA-Net w/o SepCov & 90.12 & {90.24} &95.72 &30.7\\ \hline
RAFA-Net Full Model & 91.28 &  90.47 & 96.91  & 39.1\\ 
\hline
\end{tabular}
\vspace{-0.2 cm}
\end{table} 
\vspace{-0.3 cm}
\subsection{Ablation Study}
The ablation studies imply contributions of major  modules of RAFA-Net and the results are reported in Tables \ref{tab:Abln1}-\ref{tab:AblnAug}. The  performances of key building blocks are  given in Table  \ref{tab:Abln1}. The  significance of FFN's inner components is  briefed in Table \ref{tab:Abln2}, implication  of pyramid window-size variation is reported in \ref{tab:Abln3}, and influence of other  factors  is given in \ref{tab:AblnAug}.

\subsubsection{Baseline Results} \label{BL}
To highlight the model's effectiveness for performance improvement over  bottom lines, the  baseline performances on food datasets are computed using different canonical backbone CNNs with pre-trained ImageNet weights. In this evaluation,  the output feature vector of backbone CNNs is extracted upon which a GAP layer is suited for selecting vital features. A softmax layer is added for classification in addition to a regularization layer. The  result of each  backbone is given in the first row of respective row-block of Table \ref{tab:Abln1}.

\subsubsection{Region Attention Module}
Attention focuses on features representing different complementary regions and improves learning effectiveness. This module enhances accuracy over the baselines, implying its benefits in mining descriptiveness of neighborhood regions. For example, the baseline accuracy (63.76\%) on UECF100 is substantially improved by  27.52\% using ResNet-50 base. Likewise, the gains with other backbones are justifying the  utility of this module.

\subsubsection{Feed Forward Networks}

 The discriminability could further be improved by capturing feature interactions through FFNs. To this end, the mixing of two pooling strategies enables feature selection at the pyramidal structure, and convolution enhances their interactions through FFNs  ($FFN_A$ and $FFN_B$ in Fig. \ref{fig:Model1a}  and \ref{fig:Model2a}). As a result, better performances have been obtained using FFNs over region attention, and the results are evident in Table \ref{tab:Abln1} (3rd row in each row-block). The accuracy gain of incorporating FFNs upon region attention is 1.0\% on UECF100 using ResNet-50. Similar gains of FFNs on other datasets are discriminable.    

The effectiveness of using two FFNs paths and their fusion could guide the network for further performance boost. 
The results in Table \ref{tab:Abln2} evince that accuracies have improved slightly by combining features of both FFNs than a single FFN path. It is worth  noting that without SepCov in  FFNs (``RAFA-Net w/o SepCov" in Table \ref{tab:Abln2}), the performance could decrease, implying a need of convolutional mix-up in FFNs. 
\begin{table}[!htp]
\centering
\caption{Accuracy (\%) of RAFA-Net's Various SPP Sizes  Using ResNet-50} \label{tab:Abln3}
\begin{tabular}{c c c c c c} 
 \hline
Dataset   & $SPP_{9}$   & $SPP_{14}$ & $SPP_{16}$ & $SPP_{25}$ & $SPP_{30}$ \\           \hline
UECF-100  &90.33& \textbf{91.28} & 90.02  &89.46 & 90.58\\
UECF-256  & 89.96  & \textbf{90.47} & 90.05 &89.90 &90.12 \\
MAF-121   &96.47 & \textbf{96.91} &96.08 &96.13 & 96.62\\

\hline
\end{tabular}
\vspace{-0.2 cm}
\end{table} 
\subsubsection{Weighted Feature Refinement}
The performance is boosted  by refinement  using a weighted attention technique, representing the full RAFA-Net model. The accuracy gain of weighted refinement is 0.81\% over FFNs on UECF100 using ResNet-50  (Table \ref{tab:Abln1}). Indeed, all of these key modules essentially enhance the learning and feature representation ability, evincing RAFA-Net's novelty. 
 \begin{figure*}
\centering
{\subfloat{\includegraphics[width=0.23\textwidth]{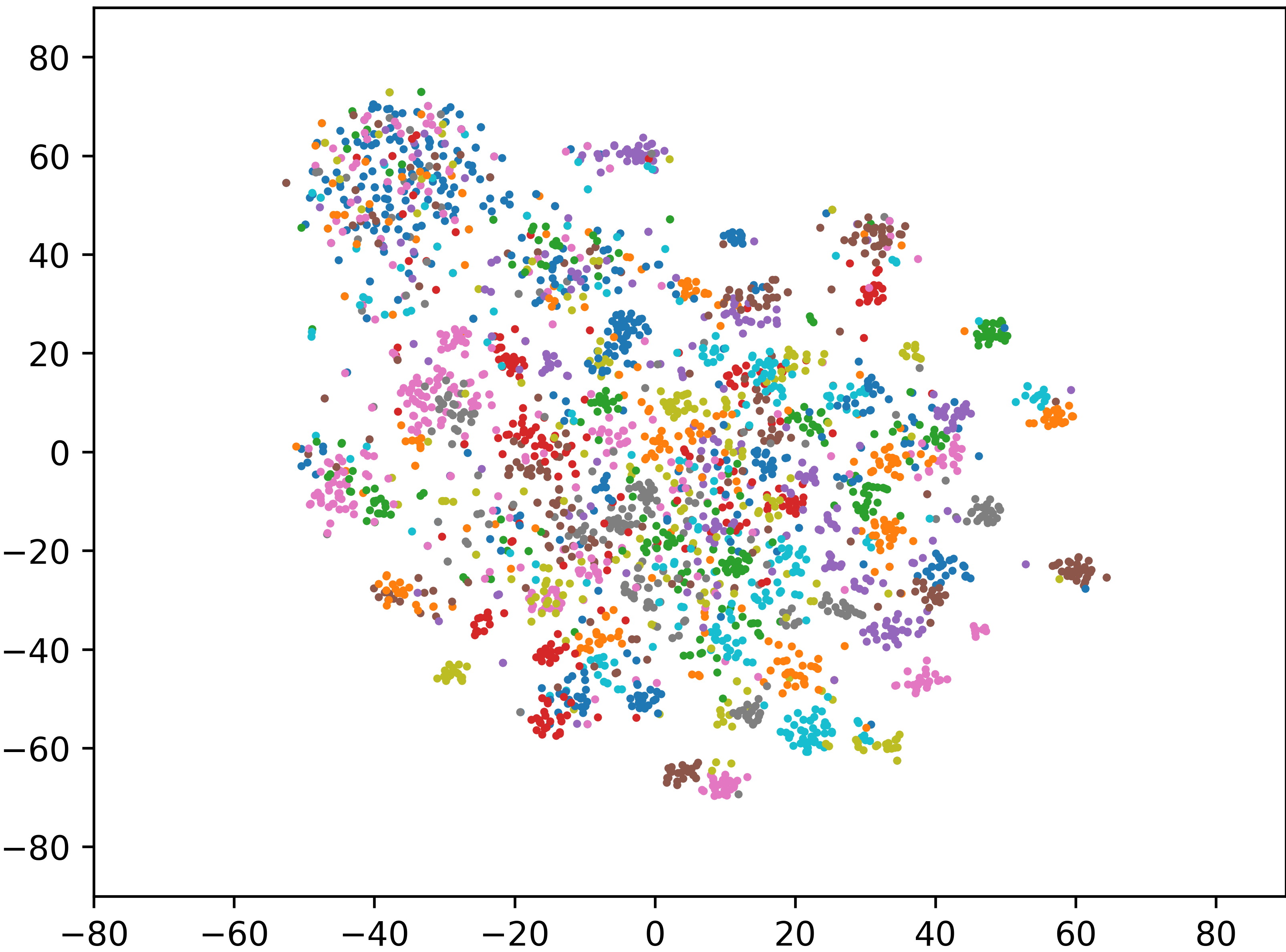}} \hfill
{\includegraphics[width=0.23\textwidth]{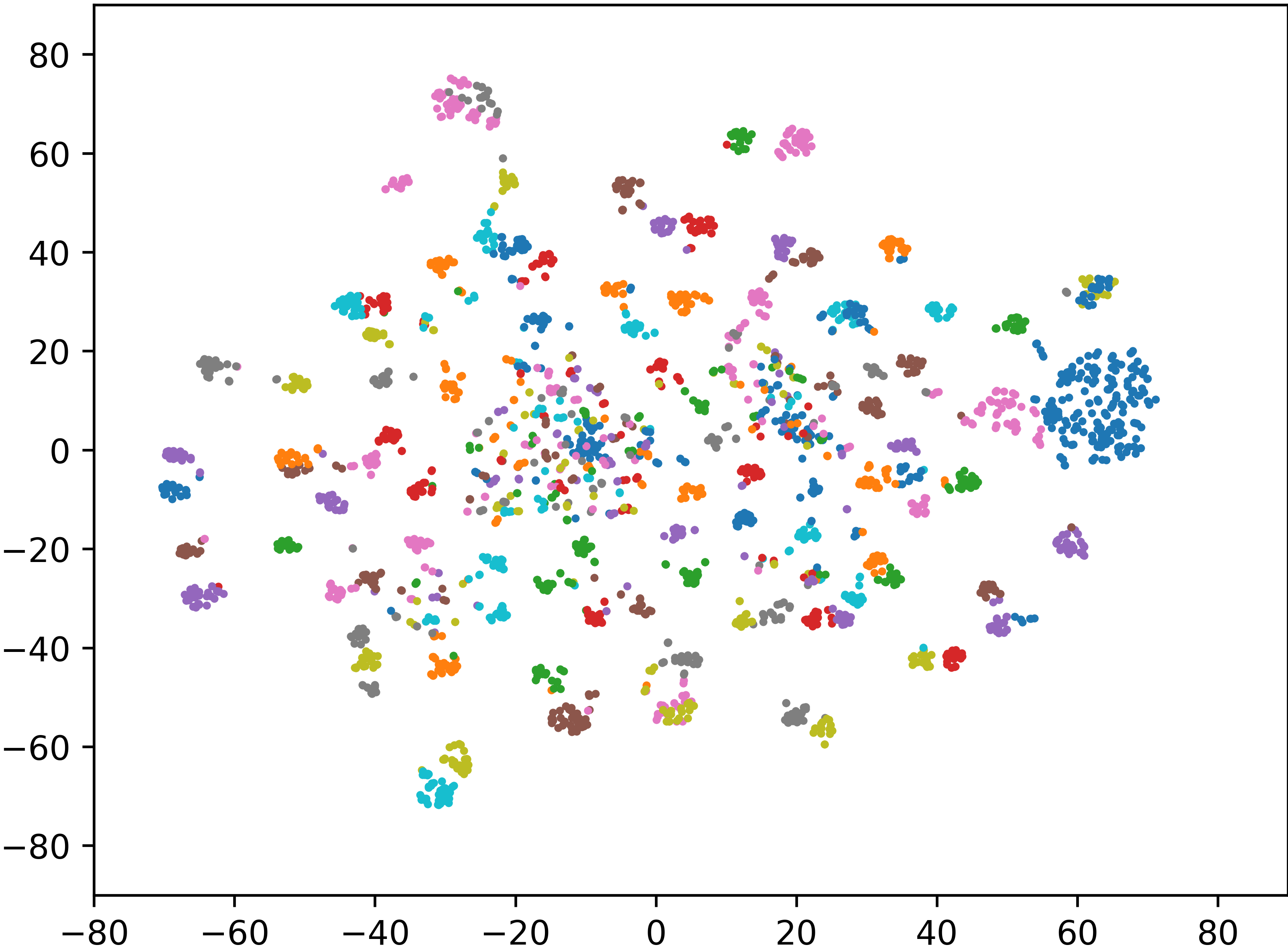}} \hfill
{\includegraphics[width=0.23\textwidth]{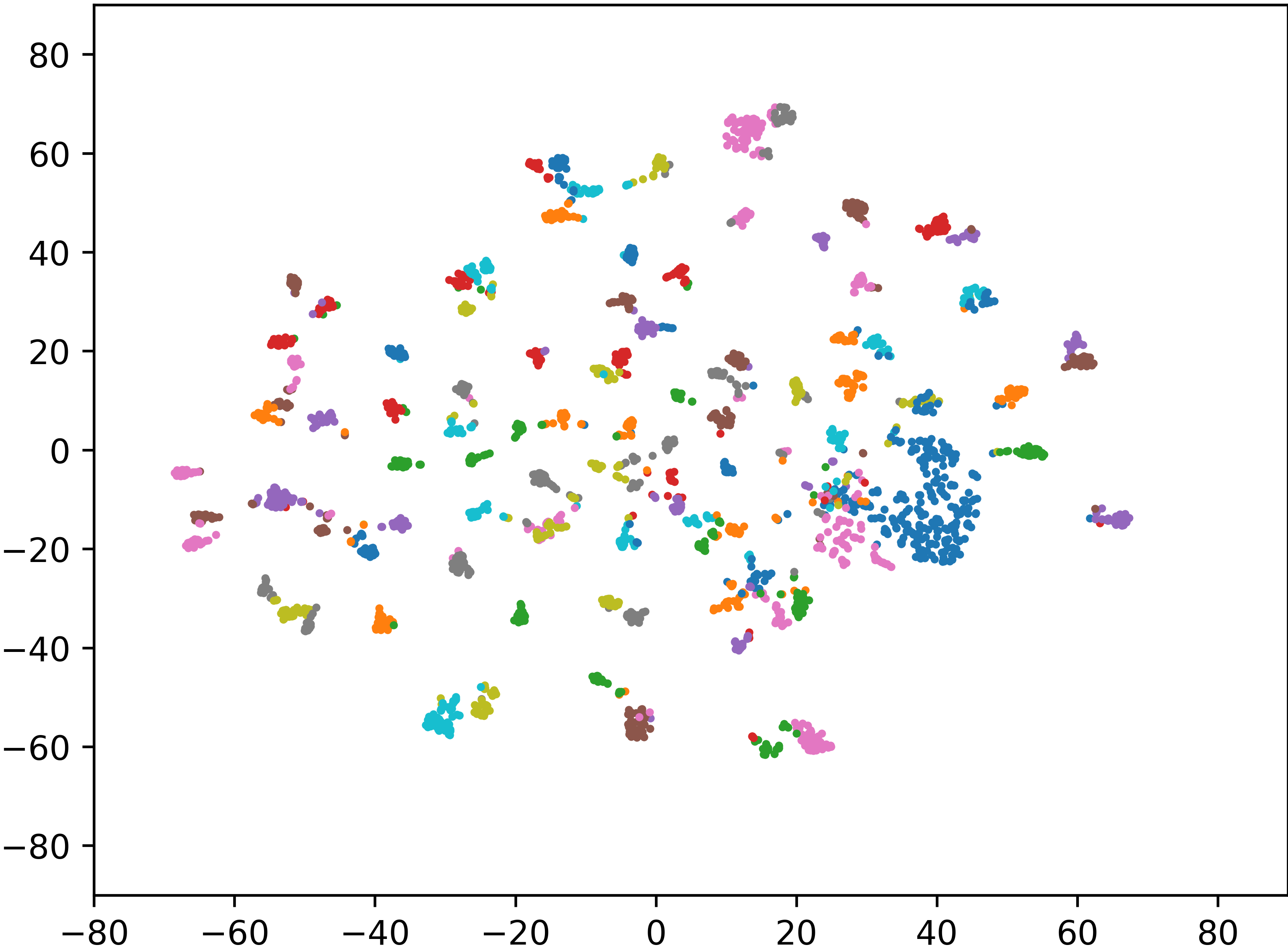}} \hfill
{\includegraphics[width=0.23\textwidth]{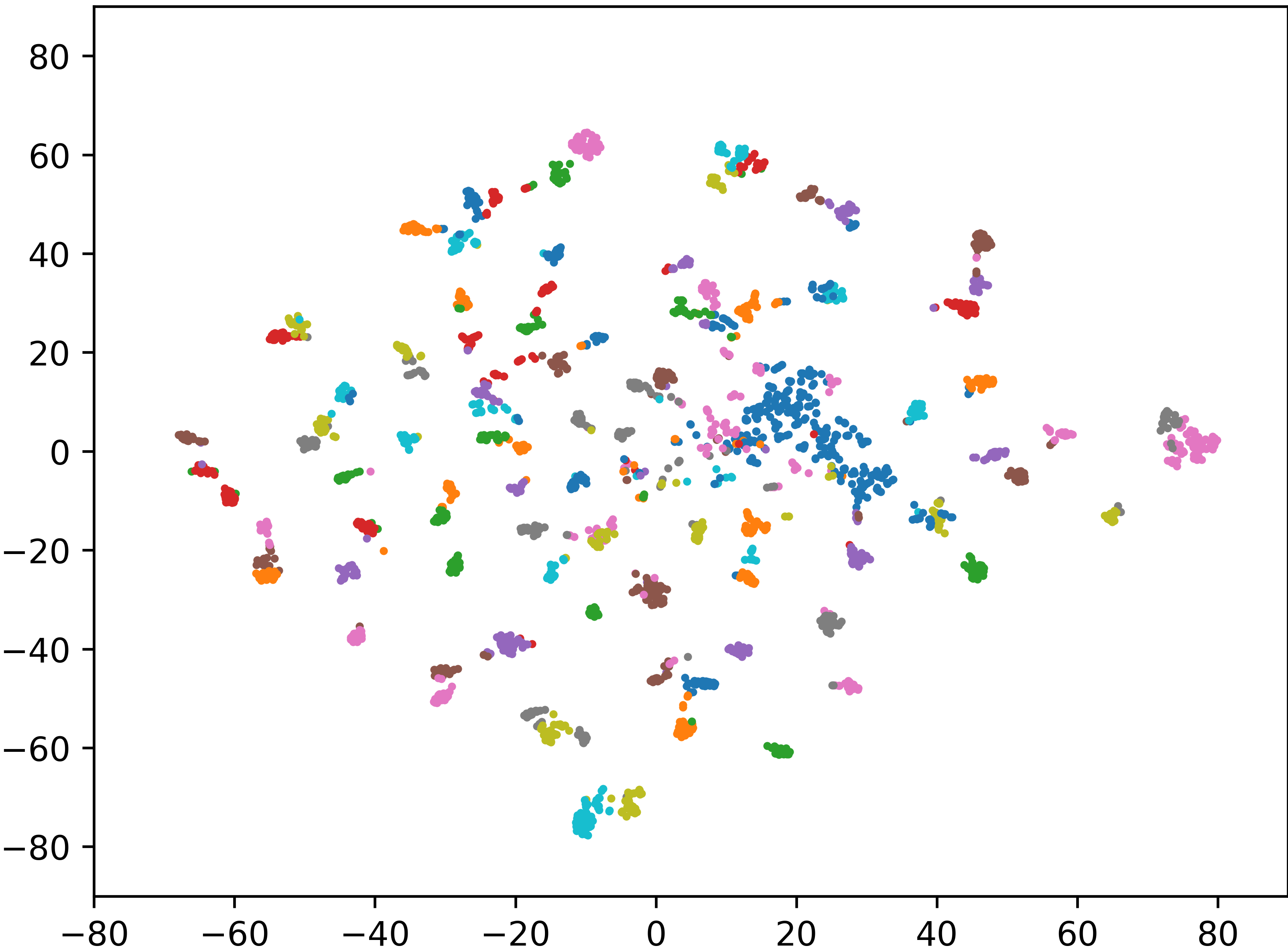}} 
}
\caption{ The t-SNE plots for feature map visualizations of RAFA-Net using ResNet-50 backbone on the UECFood-100 test dataset. Left to right: baseline, region attention, feed forward networks upon region attention, and finally, the RAFA-Net full model.}
\label{fig:tsne}
\end{figure*}
\subsubsection{Pyramid Pooling Size}
The variation of window-sizes in  spatial pyramid pooling  is tested. The results are given in  Table \ref{tab:Abln3}. The pooling pyramid with window-sizes [1$\times$1, 2$\times$2, 3$\times$3] comprising a total of 14 bins,   [3$\times$3, 4$\times$4]  for 25 bins, and finally [1$\times$1 to 4$\times$4] for 30 bins of feature vectors are used for pooling features. The results of 14 bins have improved competitively for pyramid bins of sizes 9, 16, 25, and 30. It implies that multiple window-sizes with a reasonable resolution  (\textit{e.g.}, 16$\times$16 patch-size) could represent larger spatial structures of complex  patterns.    
\begin{table}[!ht]
\centering
\caption{Accuracy(\%) of RAFA-Net Except Attention \& Other Factors } \label{tab:AblnAug}
\begin{tabular}{c c c c c} 
 \hline
Method (ResNet-50)  & UECF100  & UECF256 & MAF121  & Par (M) \\           \hline
General Data Augment  & 87.42 & 89.33 & 96.20  & 39.1\\ 
Without Self-Attention & 86.62 & 87.39 &95.55 &32.5\\
Without LayerNorm & 89.14 & 88.35 &96.32 &38.9\\
\hline
\textbf{RAFA-Net} Full Model & 91.28 & 90.47 &96.91 &39.1\\ \hline
\end{tabular}
\vspace{-0.2 cm}
\end{table}

\subsubsection{Attention and Other Important Factors}
The importance of the attention module, a mix of random occlusion and general augment, and LayerNorm are investigated. The benefits of these components of RAFA-Net are evident in performance (Table \ref{tab:AblnAug}) while evaluated excluding them. Clearly, self-attention is an effective part of  RAFA-Net, removing which accuracy reduction is obvious for all datasets. The hybrid image augment is beneficial for improving  learning effectiveness.  LayerNorm helps to boost convergence speed up and enhance performance with incurring negligible parameters in model architecture. The performance of RAFA-Net without LayerNorm (Table \ref{tab:AblnAug}) indicates an accuracy degradation compared to full RAFA-Net. Overall, the abilities of  RAFA-Net's  components are evident from the results.  

\subsubsection{Feature Map Visualization}  
The t-distributed stochastic neighbor embedding (t-SNE) \cite{van2008visualizing} visualizations exhibit the significance of different important module's layers, as shown in Fig. \ref{fig:tsne}. These figures imply that similar data points represent small pairwise distances, while different data points imply large pairwise distances using student-t distribution. Through t-SNE visual analysis, low-dimensional feature distributions of data separability in different clusters are gradually improved, showing the  discriminability  of RAFA-Net. The UECF100 test-set is used for visualizing  feature distributions. In Fig. \ref{fig:tsne}, the first image shows a canonical baseline  using ResNet-50. Second, it  shows the feature map of  region attention over base output features. Third, it  exhibits further feature enhancement  by FFNs. The last image reflects the final and full RAFA-Net. 
The variations in feature map visualizations (t-SNE clusters)  have been apprehended in performance via  ablation studies.

\subsection{Limitations,  Applications, and Future Directions}   
This work  aims to  generalize the model design for performance improvement on existing agro-food visual recognition datasets, which were collected in natural background with low image quality, depicting variations in lighting, texture, occlusion, etc. Though, the implementation of RAFA-Net based on lightweight MobileNet-v2 (param: 10M) outperforms several existing works developed using  CNNs, ViTs, ensembles, etc. yet, further improvement is necessary  for comparing with the performances attained by other deeper backbone CNN \textit{e.g.}, ResNet-50 (param: 39.1M). To reduce model complexity, a few works have tested specific crops (\textit{e.g.}, rice) by devising customized lightweight models, which might not be enough for improving performances over complex and diverse datasets (\textit{e.g.}, UECFood) grown in natural background. So, further improvement by reducing computation cost of RAFA-Net is imperative to develop a lighter DL  model for real field AI-based or robotic application. Although, existing pre-trained CNNs could be used, like our work, in embedded/edge devices due to faster convergence for training (PlantDoc:$\approx$2.7 min/epoch), and 4.3 ms/image during inference using MobileNet-v2. Currently, our method has been tested with visual images. There is an open direction for testing with other imaging modalities, \textit{e.g.}, thermal images of plant stress. 

Integration of deep learning methods with  agricultural and food processing robots in real-time is challenging for practicality. There is a rising research trend of integrating edge devices like Jetson Nano with DL models in precision agriculture \textit{e.g.}, germination analysis using segmentation \cite{donapati2023real}, weed detection, etc. Likewise, a possibility of exploiting RAFA-Net with  GPU-enabled hardware devices could be explored for realistic robotic application \textit{e.g.}, food processing, packaging, etc. The farmers would be benefited with fast analysis of plant stress  via end-user applications using a smartphone, where trained model-weight could be deployed for testing on new/unseen data. 
This type of IoT enabled framework is essential in food technology for dietary monitoring, and agriculture sector including hyperspectral imaging, drone application, etc. Sustainable  development in this domain is crucial for optimizing cost and resources in real-time deployment as hypothesized above. As deep learning  requires resource intensive framework and more training data, so, a robust solution should be developed to improve the efficacy with low-resource constraints, which is another possible future direction.

\section{Conclusion}  \label{con}
In this paper, a region attention pooling technique, dubbed RAFA-Net, is proposed for the visual categorization of food dishes and plant stress. The method integrates the distinctness of important image regions through a combination of two pooling strategies followed by an attention-guided feature refinement scheme. The RAFA-Net has boosted the performance on five benchmark datasets. An in-depth study has been provided to compare  fairly with prior works. Comprehensive ablation assessments on the main components of RAFA-Net exhibit their usefulness for model development. The  region attention module could easily be plugged into an existing  CNN. 
The RAFA-Net can be used as a cognitive and decision making module for agricultural robotics applications, \textit{e.g.}, from harvesting robots to food packaging robots. Other datasets related to agriculture and food technology  will be tested for wider applicability of  RAFA-Net for societal growth. 

\section*{Acknowledgements}
\small This work is supported by the New Faculty Seed Grant (NFSG/23-24) and Cross-Disciplinary Research Framework (CDRF: C1/23/168), and  computational infrastructure at the Birla 
Institute of Technology and Science (BITS) Pilani, Pilani Campus, Rajasthan, 333031, India. This work has been also supported in part by the project (2024/2204), Grant Agency of Excellence, University of Hradec Kralove, Faculty of Informatics and Management, Czech Republic.

\ifCLASSOPTIONcaptionsoff
  \newpage
\fi

\bibliographystyle{IEEEtran}
\bibliography{Ref.bib}
\end{document}